\documentclass{article}

\usepackage{microtype}
\usepackage{graphicx}
\usepackage{subcaption}
\usepackage{booktabs} 

\usepackage{hyperref}

\usepackage{color-edits}
\addauthor[Ellen]{ev}{blue}
\addauthor[Patrick]{py}{purple}
\addauthor[Chen]{c}{red}

\usepackage[accepted]{icml2026}

\makeatletter
\renewcommand{\printAffiliationsAndNotice}[1]{%
  \global\icml@noticeprintedtrue
}
\makeatother

\usepackage{amsmath}
\usepackage{amssymb}
\usepackage{mathtools}
\usepackage{amsthm}
\usepackage{enumitem}
\usepackage{graphicx}
\usepackage{xspace}
\newcommand{\cS}{\mathcal{S}}                   
\newcommand{\cX}{\mathcal{X}}                   
\newcommand{\cY}{\mathcal{Y}}                   
\newcommand{\R}{\mathbb{R}}

\newcommand{\E}{\mathbb{E}}

\DeclareMathOperator*{\argmax}{arg\,max}
\newcommand{\bc}{{\textsc{Bridged Clustering}}\xspace}
\usepackage{booktabs}
\usepackage{tabularx}
\usepackage{float}
\usepackage{caption}

\usepackage[capitalize,noabbrev]{cleveref}

\theoremstyle{plain}

\theoremstyle{definition}

\theoremstyle{remark}

\usepackage[textsize=tiny]{todonotes}

\icmltitlerunning{Bridged Clustering}

\begin{document}

\twocolumn[
  \icmltitle{Bridged Clustering: Semi-Supervised Sparse Bridging}



  \icmlsetsymbol{equal}{*}

\begin{icmlauthorlist}
\begin{tabular}{ccc}
\textbf{Patrick Peixuan Ye} & \textbf{Chen Shani} & \textbf{Ellen Vitercik} \\
\text{Stanford University} & \text{Stanford University} & \textbf{Stanford University} \\
\texttt{pxye@stanford.edu} & \texttt{cshani@stanford.edu} & \texttt{vitercik@stanford.edu} \\ \\
\end{tabular}
  \end{icmlauthorlist}


  \icmlcorrespondingauthor{ }{ }

  \icmlkeywords{Machine Learning, Semi-supervised Learning, Clustering}

  \vskip 0.3in
]



\printAffiliationsAndNotice{}  

\begin{abstract}
We introduce \textbf{\bc}, a semi-supervised framework to learn predictors from any \emph{unpaired} input $\mathcal{X}$ and output $\mathcal{Y}$ dataset. Our method first clusters $\mathcal{X}$ and $\mathcal{Y}$ independently, then learns a sparse, interpretable bridge between clusters using only a few paired examples. At inference, a new input $x$ is assigned to its nearest input cluster, and the centroid of the linked output cluster is returned as the prediction $\hat{y}$. Unlike traditional SSL, \bc \emph{explicitly} leverages output-only data, and unlike dense transport-based methods, it maintains a sparse and interpretable alignment. Through theoretical analysis, we show that with bounded mis-clustering and mis-bridging rates, our algorithm becomes an effective and efficient predictor. Empirically, our method is competitive with SOTA methods while remaining simple, model-agnostic, and  label-efficient in low-supervision settings.
\end{abstract}

\begin{figure*}
    \centering
    \captionsetup{justification=centering}
    \includegraphics[width=0.8\linewidth]{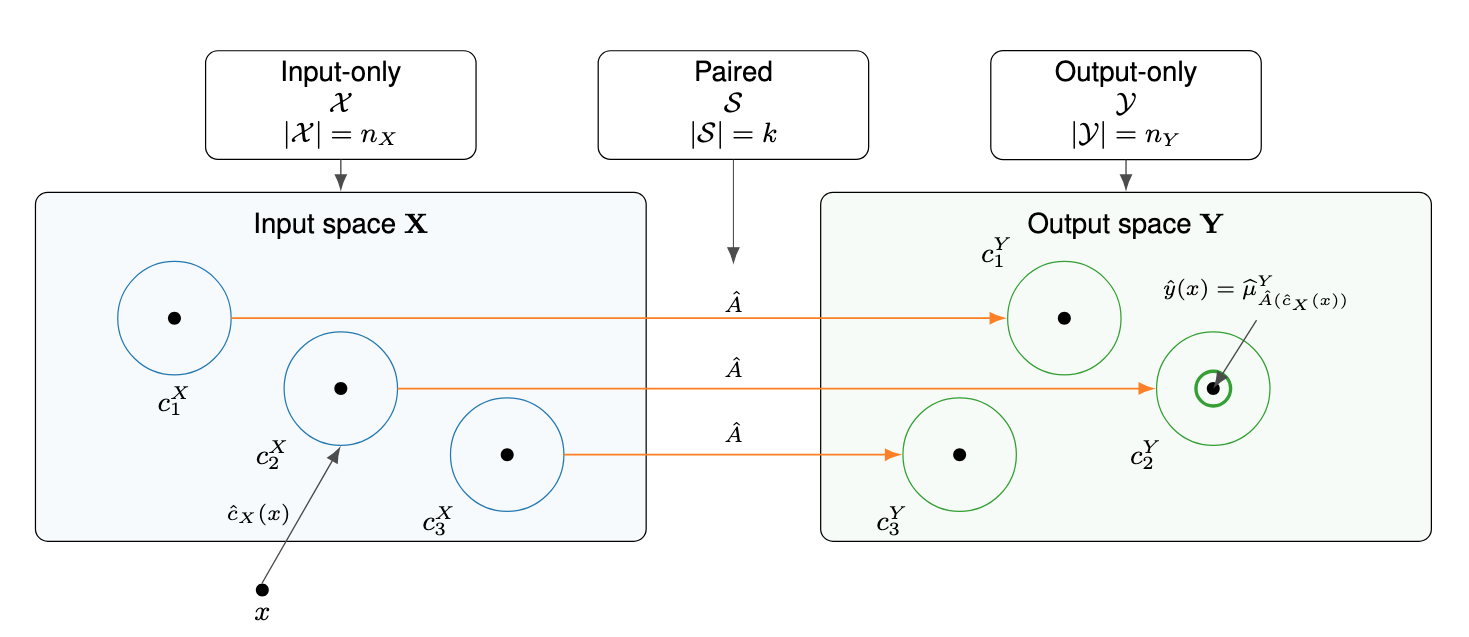}
    \caption{Bridged Clustering overview: (1) Cluster $\mathcal{X}$ and $\mathcal{Y}$ independently, (2) learn a sparse cluster bridge $\hat A$ from $k$ paired samples by majority vote, (3) predict by mapping $x$ to its input cluster and returning the centroid of the bridged output cluster.}
\end{figure*}

\section{Introduction}
\label{sec:intro}
Modern data analytics often involve reconciling rich but unaligned data collected for different purposes. For example, researchers may study a large morphological collection of museum specimen records alongside an independent genetic sequence repository such as GenBank \citep{sayers2019genbank}, without reliably coordinated specimen identifiers to join them \citep{mulcahy2022specimen}. This scenario is common across domains such as biomedicine, ecology, and the social sciences, where modern data ecosystems rarely provide perfectly paired records for every variable of interest \citep{putrama2024heterogeneous, wafa2024integration}. Building supervised models in these settings often requires costly annotation, which becomes challenging at scale.

When labeled pairs are scarce, \emph{semi-supervised learning} (SSL) is a natural approach, leveraging large pools of unlabeled inputs alongside limited supervised samples. Standard SSL assumes that each unlabeled input $x \in \mathcal{X}$ has a hidden label $y \in \mathcal{Y}$ \citep{van2020survey, yang2022survey}, but this is often violated in multimodal settings, where datasets may include \emph{output-only data} with no corresponding $x$.
Examples for output-only data include: unpaired DNA barcodes collected independently of morphological images \citep{ratnasingham2024bold}, clinical outcome records separate from patient biosignals \citep{mark2016story}, satellite imagery without matching weather station logs, audio recordings lacking textual transcriptions, or microscopy images without corresponding gene-sequencing profiles. 
In these settings, fully exploiting the available data requires learning not only from unlabeled inputs, but also from the distributional structure and abundance of unlabeled outputs. Motivated by this challenge, we ask:
\begin{quote}
    \emph{Can SSL be extended to incorporate rich, unlinked datasets of inputs and outputs for effective prediction, through simple clustering techniques?}
\end{quote}

\paragraph{Key challenges.} 
Extending SSL to settings with unlinked inputs and outputs raises several challenges. 
First, SSL objectives rely on unlabeled inputs and instance-level pairing, and therefore cannot be directly applied when outputs are observed independently of inputs.
Second, while output-only data encode valuable structural information about $\mathcal{Y}$, this information is not grounded in $\mathcal{X}$, making it unclear how it should influence prediction.
Third, approaches that align inputs and outputs only at the marginal or distributional level risk coarse solutions.
Finally, any practical method must exploit unlinked data without introducing substantial computational overhead or strong modeling assumptions.

Several lines of work partially address the challenges above.
Transport-based alignment methods exploit output-only data by matching marginal distributions across spaces, either via cross-domain costs or by aligning intra-space geometries \citep{memoli2011gromov, cuturi2013sinkhorn, peyre2019computational}. However, these approaches operate at the distributional level and may yield degenerate or overly coarse alignments.
Related work on unmatched regression leverages separate samples of inputs and outputs under restrictive structural assumptions \citep{CarpentierSchluter2016,RigolletWeed2019, AzadkiaBalabdaoui2024}.
Finally, generative and multi-view methods can effectively impute missing modalities, but typically require strong parametric assumptions or complete instance-level pairing \citep{Dempster1977, blum1998combining, bickel2004multi}. 

We address these challenges by operating at the \textit{cluster level}: we independently exploit unpaired data to structure $\mathcal{X}$ and $\mathcal{Y}$, and use a small paired set to learn a sparse, interpretable bridge between them.

{\bc} is a model-agnostic SSL framework for learning from unpaired inputs and outputs with minimal paired supervision. It clusters each space independently, learns a sparse cluster-level association from a few paired examples, and predicts by mapping inputs to their linked output clusters; exploiting structure in both modalities reduces predictive risk as unpaired data increase. We evaluate \bc across vision, language, and bioinformatics domains. Our contributions are:

\begin{itemize}[nosep]
\item We introduce a \textbf{cluster-level SSL framework} that learns from unpaired input-only and output-only data using minimal paired supervision.
\item We show how \textbf{independently collected heterogeneous datasets} can be unified through sparse cluster-level bridges, enabling effective prediction without instance-level alignment.
\item We present a \textbf{simple, interpretable, and efficient algorithm} that is model-agnostic, scales linearly in the number of examples, and yields transparent cluster-to-cluster correspondences.
\end{itemize}

Our results point to a complementary direction for SSL: \textbf{jointly exploiting input-only and output-only data through independent structure discovery}.
\footnote{We release code: \url{https://github.com/patrickpxye/bc}.}

\section{Related Work}\label{sec:related}


\paragraph{Semi-supervised and Self-supervised Learning:}

By leveraging large sets of unlabeled inputs, SSL methods such as graph-based label propagation \citep{zhu2002learning, zhou2003learning}, transductive regressors like Laplacian Regularized Least Squares (LapRLS; \citet{belkin2006manifold}), consistency regularization techniques like Mean Teacher \citep{tarvainen2017mean} and Temporal Ensembling \citep{laine2016temporal}, as well as hybrid approaches such as MixMatch \citep{berthelot2019mixmatch} and FixMatch \citep{sohn2020fixmatch} can all dramatically reduce annotation costs. For regression problems specifically, recent advances include Twin Neural Network Regression (TNNR) \citep{wetzel2022twin}, Uncertainty-Consistent Variational Model Ensembling (UCVME) \citep{dai2023semi} and and RankUp \citep{huang2024rankup}. Unlike \bc, these methods only use unlabeled inputs and do not leverage additional output-only data. Existing SSL objectives do not simultaneously consider unsupervised output alongside unlabeled inputs, even when abundant such data exist.

Self-supervised contrastive methods like SimCLR \citep{chen2020simple} involve unlabeled multi-view augmentations, and deep clustering methods such as SCAN \citep{van2020scan} and NNM \citep{dang2021nearest} also rely on multi-view consistency. These approaches use multiple augmented views, but operate within a single input modality, while our model jointly targets inputs and outputs.

\paragraph{Multi-View Learning:}
Multi-view and co-training methods explore the idea of learning from multiple distinct feature spaces, conceptually close to our setting. Related work include classical co-training and its PAC-style extensions \citep{blum1998combining, dasgupta2001pac, BalcanBlumYang2004, balcan2010discriminative}, dependency-seeking clustering approaches \citep{ReyRoth2012}, mixture-based multi-view models such as Bayesian local CCA mixtures \citep{ViinikanojaKlamiKaski2010}, and canonical least-squares clustering \citep{LeiMillerPinskyDubrawski2017}. While multi-view learning assumes co-occuring views $(X, Y)$ for each training instance, \bc solves a different problem of learning a mapping from $X$ to $Y$ where most instances are unpaired. Unlike \bc, these methods require paired views and therefore cannot use the disjoint $\mathcal{X}$- or $\mathcal{Y}$-only corpora in our problem setting.

\paragraph{Regression with Unmatched Samples:} Recent statistical work on unmatched regression address disjoint $\mathcal{X}$ and $\mathcal{Y}$ data. These work primarily rely on strong structural assumptions to recover input-output correspondence. \citet{AzadkiaBalabdaoui2024} handle linear regression as a deconvolution problem, and other related work assume monotone or isotonic structures \citep{CarpentierSchluter2016,RigolletWeed2019} to tackle unmatched regression. Another line of generative methods assumes latent parametric structure, such as mixture-of-regression optimized via Expectation–Maximization (EM) \citep{Dempster1977}. \bc makes no such assumptions that these methods are tied to, while solving the same problem for aligning disjoint inputs and outputs. 

\paragraph{Distributional alignment via Optimal Transport and GW.}
Optimal transport (OT) aligns input-output domains using a cross-space cost $c(x,y)$ and can produce predictions via \emph{barycentric mapping} \citep{cuturi2013sinkhorn,peyre2019computational}. When cross-domain metric is unavailable, \emph{Gromov–Wasserstein} (GW) matches intra-space distances to recover a soft correspondence between spaces \citep{memoli2011gromov,peyre2016gromov}. Both methods exploit output-only data through distributional structure, prone to coarse alignments and computational burdens upon scaling \citep{peyre2019computational}. Our approach differs in using a small paired set to learn a sparse cluster-bridge mapping. Whereas these methods learn dense, global couplings, our bridge is explicitly sparse, with cluster-level interpretability, and light computational overhead through only clustering and majority-voting (An asymptomatic efficiency comparison is in Appendix \ref{sec:runtime}).

\paragraph{Cross‑Domain Mapping without Parallel Data:} In NLP, unsupervised machine translation aligns monolingual embeddings and relies on back-translation to bridge languages \citep{LampleConneauDenoyerRanzato2017}. In computer vision, methods like CycleGAN learn bidirectional mappings between image domains via cycle-consistency constraints \citep{ZhuParkIsolaEfros2017,HuangLiuBelongieKautz2018}. Although these systems operate without paired instances, they train end-to-end neural generators and typically rely on reconstructive losses in both directions, while \bc is model-agnostic and learns a light cluster-level association using few labeled pairs.

\section{Notation}
\label{sec:theory}

In this section, we introduce our notation. The input features are denoted by random variable $X \in \R^d$ and the output features are denoted by random variable $Y \in \R^{d'}.$

We assume that $X$ and $Y$ are mixtures of $C$ latent groups. While the criteria defining these groups are unknown, we assume the existence of $C$ distinct categories. Let \[T\in[C]\!:=\!\{1,\dots,C\}\] be a latent categorical variable drawn with probabilities $\pi_1,\dots,\pi_C$. For each group $t$, let $P_t^X$ and $P_t^Y$ denote the distributions over $X$ and $Y$. Conditional on $T=t$, we observe a pair:
\[
  (X,Y)\;\big|\;T=t \;\sim\; P_t^{X}\times P_t^{Y},
  \]
and let $\mu_t^X := \E[X\mid T=t]$ and $
  \mu_t^Y := \E[Y\mid T=t].$

Our method takes as input the following training data:
\begin{itemize}
  \item An \emph{input pool} $\cX\!=\!\{x_i\}_{i=1}^{n_X} \subseteq \mathbb{R}^d$ of  $n_X$ i.i.d.\ draws from the marginal
        $P^X=\sum_{t}\pi_tP_t^{X}$;
  \item an \emph{output pool} $\cY\!=\!\{y_i\}_{i=1}^{n_Y}\subseteq \mathbb{R}^{d'}$of  $n_Y$ i.i.d.\ draws from the marginal $P^Y=\sum_{t}\pi_tP_t^{Y}$;
  \item a \emph{paired sample} $\cS\!=\!\{(x'_j,y'_j)\}_{j=1}^{k}$, with $k\ll n_X,n_Y$,
        drawn i.i.d. \ from joint $P=\sum_t\pi_t(P_t^{X}\!\times P_t^{Y})$.
\end{itemize}

The goal is to predict an output $Y$ from a given input $X$; the reverse direction (predicting $X$ from $Y$) is symmetric and follows by swapping roles.

\section{\bc Algorithm}\label{sec:algorithm}

\begin{algorithm}[tb]
\caption{\bc}
\label{alg:bridged_clustering}
\textbf{Input}: Input-only set $\mathcal{X} = \{ x_1, x_2, \ldots, x_{|\mathcal{X}|} \}$; Output-only set $\mathcal{Y} = \{ y_1, y_2, \ldots, y_{|\mathcal{Y}|} \}$; sparse supervised set $\mathcal{S} = \{ (x'_i, y'_i) \}_{i=1}^k$; Test-set $\mathcal{X_\text{test}} = \{ \hat{x}_1, \hat{x}_2, \ldots, \hat{x}_{|\mathcal{X_\text{test}}|} \}$\\ 
\textbf{Output}: $\hat{\mathcal{Y}} = \{ \hat{y}_1, \hat{y}_2, \ldots, \hat{y}_{|\mathcal{X_\text{test}}|} \}$
\begin{algorithmic}[1]
\STATE Apply a clustering algorithm to $\mathcal{X}$ (respectively, $\cY$) to obtain cluster assignments $\mathcal{C}_\mathcal{X}$ (respectively, $\mathcal{C}_{\cY}$).
\STATE Using the supervised set $\mathcal{S}$, learn a mapping $\hat A_{x \to y}$ between clusters in $\mathcal{X}$ and clusters in $\mathcal{Y}$.
\FOR{each sample $x_i$ in $\mathcal{X_\text{test}}$}
    \STATE Assign $x_i$ to a cluster $c_x = \mathcal{C}_\mathcal{X}(x_i)$.
    \STATE Find corresponding cluster in $\mathcal{Y}$: $c_y = \hat A_{x \to y}(c_x)$.
    \STATE Predict $\hat{y}_i$ as the centroid of cluster $c_y$ in $\mathcal{Y}$.
\ENDFOR
\STATE \textbf{return} Predicted outputs $\hat{\mathcal{Y}}$ for test inputs $\mathcal{X_\text{test}}$
\end{algorithmic}
\end{algorithm}

\bc (Algorithm~\ref{alg:bridged_clustering}), first independently clusters the input space $\mathcal{X}$ and the output space $\mathcal{Y}$. Next, it learns a bridge mapping $\hat{A}$ from the paired samples, where each input cluster is matched to an output cluster via majority vote. Finally, for inference, a new input $x$ is assigned to its nearest input cluster, and the prediction $\hat{y}$ is given by the centroid of the linked output cluster. 

\paragraph{Cluster assignments:} We first run a $k$-clustering algorithm to partition the dataset into $k$ groups separately on $\cX$ and $\cY$ using $C$ centers:
\[
  \hat c_X:\R^d\!\to[C],
  \hat c_Y:\R^{d'}\!\to[C].
\]
Because cluster labels are only defined up to a permutation of
indices, we evaluate their quality via the permutation-invariant
mis-clustering rates:
\[
  \varepsilon_X := \min_{\sigma_X:[C]\!\to\![C]}\Pr\bigl[\sigma_X(\hat c_X(X))\neq T\bigr]
\]
\[
  \varepsilon_Y := \min_{\sigma_Y:[C]\!\to\![C]}\Pr\bigl[\sigma_Y(\hat c_Y(Y))\neq T\bigr].
\]
Let \(\sigma_X,\sigma_Y\) denote permutations that attain these minima, and let $\sigma_X^{-1},\sigma_Y^{-1}$ denote their inverses.

\paragraph{Bridging from few pairs:} On the $k$ paired points, we \emph{match} input-space clusters to output-space clusters by majority vote. For every input cluster $a \in [C]$, mapping \[\hat A(a) \;:=\;
  \argmax_{b\in[C]}
  \sum_{(x',y')\in\cS} \mathbf{1}\!\{\hat c_X(x')=a,\;\hat c_Y(y')=b\}.\]
For a random latent label $T$, we call
\[
\varepsilon_B := \Pr\!\left[\sigma_Y\!\left(\hat A(\sigma_X^{-1}(T))\right)\neq T\right].
\] the \emph{mis-bridging rate}.
Algorithm~\ref{alg:bridged_clustering}
predicts a fresh output by returning the empirical centroid of the
output cluster indicated by $\hat A\circ\hat c_X$. Let \(\widehat\mu_b^Y\) be the empirical mean of \(\cY\)-points assigned to cluster \(b\), so our final predictor is \[\hat Y(x) := \widehat{\mu}_{\hat A(\hat c_X(x))}^{\,Y}.\]

\section{Experiments}
\label{sec:experiments}

We now turn to empirically test the predictive accuracy of \bc. We use fully paired datasets with known input-output correspondences ($(x_i, y_i) \in \mathcal{X} \times \mathcal{Y}$) across various modalities. These datasets allow us to simulate independently collected $\mathcal{X}$-only and $\mathcal{Y}$-only repositories by hiding most of the pairings and exposing only a small, randomly selected subset during training. This controlled setup with ground-truth pairs enables us to measure performance as a function of the number of labeled pairs. 

\subsection{Data} We apply \bc on four multimodal datasets that connect distinct feature spaces: the \emph{BIOSCAN-5M} dataset that pairs specimen images with DNA barcode sequences of insect samples \citep{gharaee2024bioscan}, the \emph{WIT} dataset \citep{srinivasan2021wit} pairing images with their text captions on Wikipedia pages, as well as the \emph{Flickr30k} dataset \citep{young2014image} which connects images of people engaged in everyday activities with human-annotated captions and the \emph{COCO} dataset \citep{lin2014microsoft} that connects common object images with text descriptions.

For experimental evaluation, we randomly sample a number of \textit{data groups}, where each group corresponds to a coherent semantic category or label within the dataset (e.g., an insect species in BIOSCAN, or an object category in COCO). Each run includes a subset of these groups (e.g., 5 species of insects from BIOSCAN or 7 categories of objects from COCO) to assess \bc's ability to recover and align latent structure. We detail the data, processing, and mapping in Appendix \ref{app:details}.

\subsection{Experimental Overview} We evaluate \bc's ability to predict the output features $\mathcal{Y}$ (e.g., DNA sequences or text captions) from the input features $\mathcal{X}$ (e.g., images). Since \bc is bidirectional, we also test the inverse by predicting input features $\mathcal{X}$ from output features $\mathcal{Y}$. 

For each dataset, we simulate a disjoint data setting by designating a small subset of datapoints as the supervised set, which contains paired input-output examples. The remaining datapoints are divided into two unpaired sets: one with input-only features and the other with output-only features, mimicking independent data sources. We run experiments in two different settings: transductive, where the model has access to the unlabeled test inputs during training (with test outputs withheld), and inductive, where evaluation is performed on entirely unseen test data.

\noindent \textbf{Metrics.} We compute the Mean Squared Error (MSE) between the predicted and the ground-truth output features. For our BIOSCAN experiment, we directly measure the MSE between the algorithm-predicted embedding vector and the ground truth DNA embedding vector. In all three image captioning tasks (WIT/Flickr30k/COCO), we consider an additional metric that directly measures the accuracy of the predicted caption: BLEU score, besides the MSE. Since our outputs are text embeddings, we map each predicted embedding to its nearest neighbor in the available output set, which is a text embedding that corresponds to some actual text. We retrieve this text as our predicted caption. We then compute the BLEU score between this retrieved caption and the ground-truth caption. (see details in Appendix~\ref{app:details}).

\subsection{Baselines} We compare \bc against a comprehensive suite of baselines spanning supervised learning, classical transductive methods, graph-based models, consistency-regularized deep learners, pairwise regression approaches, and unmatched regression techniques (see more information in Section \ref{sec:related}). Together, they span the full landscape of SSL and related paradigms, whereas all except KNN leverage unlabeled inputs and paired data, and \bc, KMM, EM, EOT, GW incorporate unlabeled outputs as well. The following list categorizes the baselines we compare \bc against:
\begin{itemize}[nosep, leftmargin=5.5mm]
    \item \textbf{Supervised:} K-nearest neighbors (KNN).
    \item \textbf{Transductive SSL:} Laplacian Regularized Least Squares (LapRLS) and Transductive SVR (TSVR; the standard Support-Vector Regression was adapted to semi-supervised setting).
    \item \textbf{Graph-based:} Graph Convolutional Network (GCN).
    \item \textbf{Consistency-regularized:} FixMatch.
    \item \textbf{Low-supervision regressors: }Twin Neural Network Regression (TNNR), and Uncertainty-Consistent Variational Model Ensembling (UCVME).
    \item \textbf{Unmatched-regression:} Kernel Mean Matching (KMM), Mixture-of-Regressions via EM.
    \item \textbf{Transport-based:} Entropic OT with Barycentric Mapping (EOT), Gromov-Wasserstein (GW).
    \end{itemize}
\subsection{Results}

\begin{table*}[t]
\centering
\caption{Win-rates across transductive (a) and inductive (b) experiments. Bold indicates the best model per dataset (each row is summed to 100\%). Win-rates are defined as the model with the lowest MSE across 600 runs in every dataset (30 randomized trials per setting).}
\setlength\tabcolsep{2pt} 
\textbf{(a) Transductive experiments} \\[0.3em]
\begin{tabular}{l|c|c|c|c|c|c|c|c|c|c|c|c}
\toprule
Exp. & BC & KNN & FM & LapR. & TSVR & TNNR & UCVME & GCN & KMM & EM & EOT & GW \\
\hline
BIOSCAN              & \textbf{67\%} & 0\% & 0\%  & 0\% & 19\% & 0\% & 0\% & 0\% & 13\% & 1\% & 1\% & 0\% \\
BIOSCAN (rev.)       & \textbf{63\%} & 0\% & 0\% & 0\% & 0\% & 0\% & 0\% & 0\% & 1\% & 1\% & 36\% & 0\% \\
COCO                 & \textbf{71\%} & 0\% & 0\% & 0\% & 0\% & 0\% & 0\% & 0\% & 0\% & 3\% & 26\% & 0\% \\
COCO (rev.)          & \textbf{89\%} & 0\% & 0\% & 0\% & 0\% & 0\% & 0\% & 0\% & 0\% & 4\% & 7\% & 0\% \\
Flickr30k            & \textbf{56\%} & 0\% & 0\% & 0\% & 0\% & 4\% & 0\% & 0\% & 1\% & 2\% & 36\% & 1\% \\
Flickr30k (rev.)     & \textbf{71\%} & 0\% & 0\% & 0\% & 13\% & 0\% & 0\% & 0\% & 3\% & 0\% & 10\% & 4\% \\
WIT                  & 12\% & 0\% & 0\% & 0\% & 0\% & \textbf{32\%} & 0\% & 0\% & 28\% & 3\% & 16\% & 10\% \\
WIT (rev.)           & 9\% & 0\% & 0\% & 0\% & 4\% & \textbf{30\%} & 0\% & 0\% & 27\% & 0\% & 5\% & 25\% \\
\bottomrule
\end{tabular}

\vspace{0.7em} 

\textbf{(b) Inductive experiments} \\[0.3em]
\begin{tabular}{l|c|c|c|c|c|c|c|c|c|c|c|c}
\toprule
Exp. & BC & KNN & FM & LapR. & TSVR & TNNR & UCVME & GCN & KMM & EM & EOT & GW \\
\hline
BIOSCAN              & \textbf{67\%} & 0\% & 0\% & 0\% & 18\% & 0\% & 0\% & 0\% & 11\% & 1\% & 3\% & 0\% \\
BIOSCAN (rev.)       & \textbf{61\%} & 0\% & 0\% & 0\% & 0\% & 0\% & 0\% & 0\% & 1\% & 1\% & 37\% & 0\% \\
COCO                 & \textbf{70\%} & 0\% & 0\% & 0\% & 0\% & 0\% & 0\% & 0\% & 0\% & 3\% & 28\% & 0\% \\
COCO (rev.)          & \textbf{84\%} & 0\% & 0\% & 0\% & 0\% & 0\% & 0\% & 0\% & 0\% & 5\% & 11\% & 0\% \\
Flickr30k            & 45\% & 1\% & 0\% & 0\% & 0\% & 4\% & 0\% & 0\% & 1\% & 3\% & \textbf{46\%} & 1\% \\
Flickr30k (rev.)     & \textbf{60\%} & 1\% & 0\% & 0\% & 0\% & 1\% & 0\% & 0\% & 3\% & 1\% & 32\% & 3\% \\
WIT                  & 12\% & 0\% & 0\% & 0\% & 0\% & \textbf{32\%} & 0\% & 0\% & 28\% & 3\% & 16\% & 10\% \\
WIT (rev.)           & 9\% & 0\% & 0\% & 0\% & 4\% & \textbf{30\%} & 0\% & 0\% & 27\% & 0\% & 5\% & 25\% \\
\bottomrule
\end{tabular}
\label{tab:combined}
\end{table*}

Each run of our experiments samples a subset of groups (e.g., species in \textsc{BIOSCAN}) from our datasets. We test the robustness of \bc by sampling different numbers of data groups $\in \{3, 4, 5, 6, 7\}$ and varying the numbers of supervised samples per cluster $\in \{1, 2, 3, 4\}$\footnote{Corresponds to $\{0.5\%, 1\%, 1.5\%, 2\%\}$ supervision for BIOSCAN and COCO (cluster size=200), and $\{4\%, 8\%, 12\%, 16\%\}$ for WIT and Flickr30k (cluster size=25).}, yielding $4\times 5 = 20$ different settings per experiment.

Each setting is run 30 times with different seeds. Table~\ref{tab:combined} reports win-rates (lowest MSE across 600 runs per dataset), and Figure~\ref{fig:res_main_paper} shows MSE distributions. Additional results in Appendix Figures \ref{fig:figa}-\ref{fig:figf}. Overall, \textbf{\bc outperforms all baselines across all datasets except WIT}. We note the robustness of the results across transductive and inductive setups, and the fact that \bc is always better in the transductive configuration.

\begin{figure*}[h!]
    \centering
    \includegraphics[width=0.9\linewidth]{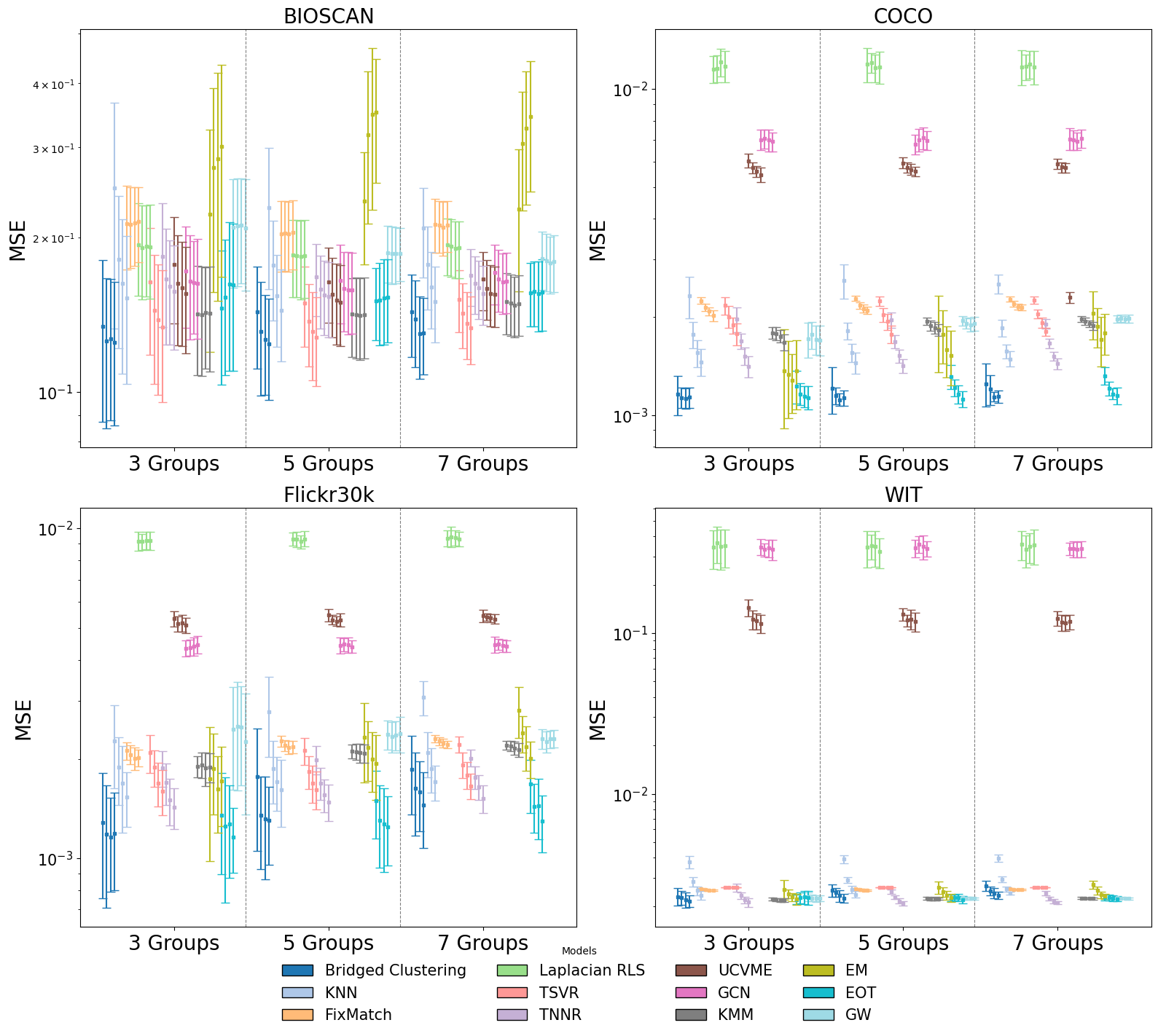} 
    \caption{MSE distribution of different models in the inductive setting. The four distribution plots of the same color represent the settings with 1, 2, 3, and 4 supervised samples per cluster. See more detailed MSE figures in the Appendix \ref{app:mse}.}
    \label{fig:res_main_paper}
\end{figure*}

\subsection{Discussion}
\bc outperforms baseline performance in three out of four datasets. In these three datasets (we will discuss WIT's performance shortly), Wilcoxon signed-rank tests confirmed these findings are statistically significant (largest p-value $=6\mathrm{e}{-4}$).

\paragraph{\textbf{Robust success:}}Across both directions ($\mathcal{X}\rightarrow\mathcal{Y}$, $\mathcal{Y}\rightarrow\mathcal{X}$) in BIOSCAN, Flickr30k, and COCO, \bc achieves consistently low MSE, \textbf{outperforming all baselines over varied cluster sizes and supervision levels}. In WIT, it falls short of the best model, but remains highly competitive, notably surpassing all SSL baselines by leveraging output-only data $\mathcal{Y}$. On average, \bc also outperforms KMM, EM, EOT, and GW, despite these methods using full training data, underscoring the strength of our simpler approach. Moreover, unlike transport-based methods, \bc does not require solving costly optimal transport problems, leading to lower computational overhead, and produces easily interpretable cluster assignments aligned with conceptual groupings in the data (See Illustrations in Appendix \ref{app:illu}).

\paragraph{\textbf{Cluster quality and accuracy:}} {\bc}'s performance depends on two factors: recovering latent structure via independent clustering (cluster quality) and aligning clusters with limited supervision (bridging accuracy). We observe that \textbf{cluster quality is negatively correlated with model loss}, measured using adjusted mutual information (AMI) (Figure A.\ref{fig:fig2} in Appendix \ref{app:cluster_quality}). WIT's low mutual information scores suggests that weaker clustering explains {\bc}'s lower performance on this dataset. We also observe a positive correlation between bridging accuracy and cluster quality in \ref{fig:figh} and \ref{fig:figi}, indicating that these two performance-relevant factors are correlated, a connection that'll be rediscussed in Section \ref{sec:thedis}.

\paragraph{Low supervision regime:} \bc performs well in low-supervision settings, as it requires as few as one labeled example per cluster to associate input and output clusters accurately. While increasing the number of supervised points can improve alignment accuracy, it has a limited effect on prediction quality. We can see this in Appendix \ref{app:bleu}, where Figures \ref{fig:figv} - \ref{fig:figx} show that lower supervision (1,2 sample) yields more competitive performance then higher supervision (3,4 sample) when compared to other methods, as measured by the task accuracy metric BLEU. In other words, since the model always returns the output cluster centroid, further supervision does not refine individual predictions. In contrast, other models may benefit more from additional supervision. To address the potential need for cluster-wise refinement, we designed an algorithmic variant in Section \ref{clusterwise}. 

\paragraph{Transductive Versus Inductive:} The performance of Bridged Clustering across both transductive and inductive settings are consistently competitive, with slightly more wins in the transductive tests. Given that our clustering-based method naturally takes advantage of inference points to better learn the cluster geometry in the transductive settings, this marginal win is expected.

\paragraph{\textbf{Higher risk with more data groups:}} 
\bc's relative weakness in high-cluster regimes stems from its reliance on bijection input-output mapping. As cluster count increases, the space of possible mappings grows exponentially, making accurate alignment harder (see Figure \ref{fig:bridg_acc} in Appendix \ref{app:cluster_quality}). Baseline methods without such explicit mappings prove more robust to increases in cluster count, highlighting \bc's tradeoff: it \textbf{excels in low-supervision, low-cluster settings, but becomes more sensitive as structural complexity increases}.

\section{Algorithmic Variants and Extended Applicability}

To preserve the conceptual simplicity of our exposition, our main \cref{alg:bridged_clustering} adopts general primitives: any clustering method for unsupervised learning and any mapping mechanism for sparse bridging.

To illustrate how \bc could be adapted to more complex settings, we analyze two algorithmic variants, each addressing a range of data-centric challenges. In \cref{sec:soft,clusterwise}, we describe these variants, and in \cref{sec:variant_experiments}, we describe the main takeaways of our empirical results using these variants under challenging data conditions.

\subsection{Soft Bridging for Flexible Mapping}\label{sec:soft}

When applying our algorithm to datasets without clearly separated categories, clustering primitives may become prone to overlapping or miscounted clusters. Rigid one-to-one mapping, as in \cref{sec:algorithm}, becomes vulnerable to misattributed clusters under this setting.

Hence, we design a soft-bridging function as our new mapping primitive: using a small supervised set to build input-output cluster bridges, we construct probabilistic bridges between every possible input-output cluster pair, dynamically weighting each edge by the number of supervised points connecting the clusters. Formally, let $w_{i \to j}$ denote the fraction of supervised points whose inputs fall in input cluster $i$ and whose labels lie in output cluster $j$, with $\sum_{j} w_{i \to j} = 1$.  For an input $x$ assigned to cluster $i$, soft bridging predicts a weighted combination of all output cluster centroids:
\[
    \hat y(x)
    \;=\;
    \sum_{j=1}^{C} w_{i \to j}\,\bar y_j .
\]
This method reduces the impact of enforcing a rigid one-to-one bridge when the clustering is imperfect. In practice, the clustering step may return the wrong number of clusters or be unable to distinguish overlapping clusters, which can lead to incorrect cluster correspondences. Soft bridging avoids committing to a single correspondence by distributing probability mass across output clusters, and our experiments (\cref{sec:variant_experiments}) show that it is robust on noisy data.

\subsection{Cluster-wise Regression For Refined Prediction}
\label{clusterwise}
Once a test point $x$ is assigned to an input cluster $i$ and bridged to an output cluster $j$, \cref{alg:bridged_clustering} returns the output centroid $\widehat{\mu}_j^Y$, collapsing intra-cluster variation. To relax this rigidity, we propose cluster-wise regression, deployed as a third step after clustering and bridging. For each cluster pair, we use in-cluster supervised points to fit a linear map $f_{i \to j}(x) \in \mathbb{R}^{d'}$  and predict
\[
    \hat y(x)
    \;=\;
    (1 - \alpha)\,\widehat{\mu}_j^Y
    \;+\;
    \alpha\, f_{i \to j}(x),
\]
with $\alpha \in [0,1]$ controlling the strength of the regression refinement. We instantiate two variants:
\begin{itemize}[leftmargin=*, itemsep=1pt]
    \item \textbf{\textsc{Supervised Refinement}}: linear regression on the supervised pairs in $(i,j)$;
    \item \textbf{\textsc{Supervised+Centroid Refinement}}: linear regression on the supervised pairs \emph{plus} the $C$ estimated input/output cluster centroids.
\end{itemize}

Cluster-wise regression is especially valuable as template for extending \bc\ beyond very-low-supervision and conditionally independent settings. In regimes where the output $Y$ is conditionally linearly on the input $X$ given the clustered latent $T$, this offers a direct way to refine our cluster-based predictor, if provided sufficient labeled pairs to fit these cluster-wise regressors. Concretely, for a given $T = t$, if the relationship between $X$ and $Y$ is well approximated by a cluster-wise linear map $Y \approx W_t X + b_t$, we can reliably estimate that map with sufficient labeled pairs within each cluster.

\subsection{Experiments with Challenging Data Conditions}\label{sec:variant_experiments}

In order to target \emph{overlapping} clusters, we design a new experiment by sampling multi-labeled COCO classes that share at least one overlapping label with another sampled class, in contrast to our earlier use of single-labeled categories. Meanwhile, in order to target a \emph{misspecified number} of clusters, we design another experiment in COCO by intentionally mis-specifying the number of clusters in our K-clustering primitives.

In the two experiments, both Soft-Bridging and Cluster-wise Regression (Supervised+Centroid Refinement) outperform \cref{alg:bridged_clustering} under these challenging data settings, suggesting the adaptivity of \bc as an effective template for diverse algorithms that jointly exploit unsupervised inputs and outputs. 
See Figure \ref{fig:figk} through \ref{fig:fign} in Appendix \ref{app:variant} and \ref{app:adver}.

\section{Theoretical Discussion}
\label{sec:thedis}

This section provides the mathematical principals behind \bc, identifying key factors governing its effectiveness. We also provide a runtime analysis in Appendix ~\ref{sec:runtime}.

Representation quality and scarce supervision influence the final accuracy of \bc through several key sources of error, including the input mis-clustering $(\varepsilon_X)$, output mis-clustering $(\varepsilon_Y)$, and mis-bridging $(\varepsilon_B)$. The main takeaways of this section are that (1) stronger embeddings that enlarge inter-clustering separation directly reduce $\varepsilon_X$ and $\varepsilon_Y$, and (2) once clusters are reliable, even a small handful of paired examples suffice to learn the sparse cluster bridge with exponentially small error.

\paragraph{Mis-clustering under sub-Gaussian mixtures.} Suppose the input and output distributions are $\sigma^2$–sub-Gaussian with means $\mu_t^X,\mu_t^Y$. Define the minimum separation \[\Delta_X = \min_{t \ne s \in [C]} \|\mu_t^X - \mu_s^X\|,\Delta_Y = \min_{t \ne s \in [C]} \|\mu_t^Y - \mu_s^Y\|.\] It is known that under Lloyd's algorithm, under good initialization, $C$-means clustering on $n_X$ unlabeled inputs satisfies
\[
\varepsilon_X \le \exp\!\left(-\frac{\Delta_X^2}{16\sigma^2}\right)
\quad \text{for all } s \ge 4\log n_X
\]
with probability at least $1 - \nu - \frac{4}{n_X} - 2\exp(-\Delta_X/\sigma)$, where $\nu$ is the initializer failure rate~\citep{lu2016statistical}. An identical bound holds for $\varepsilon_Y.$ Thus, any improvement in embeddings that increases $\Delta_X$ or $\Delta_Y$ sharpens the bound without changing the algorithm. Consistent with this theory, we observe a strong negative correlation between cluster quality (AMI) and predictive loss (Appendix~\ref{app:cluster_quality}).

\paragraph{Efficient bridging from few pairs.} Once the mis-clustering rates $\varepsilon_X, \varepsilon_Y$ are controlled, learning the bridge becomes simple. As observed in our experiments, there's a positive correlation between bridging accuracy and cluster quality. For a supervised pair $(X,Y)|T \in \mathcal{S}$, consider the events $\{\sigma_X(\hat c_X(X)) = T\}$ and $\{\sigma_Y(\hat c_Y(Y)) = T\}$, which correspond to correct cluster assignments in the input and output spaces, controlled via $\varepsilon_X, \varepsilon_Y$. If both events hold with sufficiently high probability $(>50\%)$, then a majority vote over just a few labeled pairs will suffice to recover the correct cluster-to-cluster assignment, resulting in a low mis-bridging error $\varepsilon_B$.
Empirically, we find that even one or two examples per cluster are enough for reliable bridging, explaining why \bc remains effective with less than \(1\%\) paired supervision (\cref{sec:experiments}).

\section{Conclusion \& Future Work}\label{sec:conclusions}

\bc offers a simple new perspective for SSL by elevating \emph{output-only} data from a peripheral role to a central asset. By decoupling input and output representations and connecting them via a sparse, interpretable bridge, \textbf{our framework addresses a gap between classic input-centric SSL and modern data ecosystems with abundant but misaligned observations}.

Empirically, \bc performs competitively across diverse domains and modalities. The method is model-agnostic, computationally efficient, and supports bidirectional inference without retraining. By separating representation learning from supervision, it uncovers latent structure in both input and output spaces and aligns them with minimal labels. \textbf{Concept-level alignment between independently clustered spaces provides a strong inductive bias, enabling generalization without dense supervision}, which is especially valuable in settings with limited labeled data, high domain variability, or distribution shifts.

Our framework is naturally extensible while preserving interpretability and efficiency. Extending our model to algorithmic variants such as Soft-Bridging and Cluster-wise Regression, brings robustness in adversarial data conditions. Future work can explore combination of \bc with unsupervised techniques like manifold learning or self-supervised embeddings, in effort to build more flexible semi-supervised learning systems capable of leveraging richly available but misaligned data.

Our findings suggest a future where modular, concept-aligned learning can extract meaningful structure from unsupervised but richly available observations, without the need for dense labels or heavy models.

\section*{Impact Statement}

\bc primarily advances methodology in machine learning and does not directly involve human subjects or sensitive personal data. Nonetheless, we acknowledge that methods for automated clustering and prediction can be applied in real-world settings with potential social impact.

The design of \bc emphasizes interpretability and low supervision, which can help mitigate risks associated with opaque or biased models. By explicitly linking clusters and maintaining cluster-level transparency, users can inspect and understand the learned correspondences, reducing the likelihood of unintended harm from unexamined correlations.

We caution that, like any predictive system, applying \bc to high-stakes domains (e.g., hiring, law enforcement, healthcare) requires careful consideration of fairness, privacy, and potential biases in the data. Our experiments are conducted on publicly available benchmark datasets, and no sensitive or personally identifiable information is used.

\bibliography{mainbib}

@article{belkin2006manifold,
  title={Manifold regularization: A geometric framework for learning from labeled and unlabeled examples.},
  author={Belkin, Mikhail and Niyogi, Partha and Sindhwani, Vikas},
  journal={Journal of machine learning research},
  volume={7},
  number={11},
  year={2006}
}

@misc{zhu2002learning,
  title={Learning from labeled and unlabeled data with label propagation.(2002)},
  author={Zhu, Xiaojin and Ghahramani, Zoubin},
  year={2002}
}

@inproceedings{chen2020simple,
  title={A simple framework for contrastive learning of visual representations},
  author={Chen, Ting and Kornblith, Simon and Norouzi, Mohammad and Hinton, Geoffrey},
  booktitle={International conference on machine learning},
  pages={1597--1607},
  year={2020},
  organization={PmLR}
}

@inproceedings{van2020scan,
  title={Scan: Learning to classify images without labels},
  author={Van Gansbeke, Wouter and Vandenhende, Simon and Georgoulis, Stamatios and Proesmans, Marc and Van Gool, Luc},
  booktitle={European conference on computer vision},
  pages={268--285},
  year={2020},
  organization={Springer}
}

@article{mark2016story,
  title={The story of MIMIC},
  author={Mark, Roger},
  journal={Secondary Analysis of Electronic Health Records},
  pages={43--49},
  year={2016},
  publisher={Springer}
}

@inproceedings{dang2021nearest,
  title={Nearest neighbor matching for deep clustering},
  author={Dang, Zhiyuan and Deng, Cheng and Yang, Xu and Wei, Kun and Huang, Heng},
  booktitle={Proceedings of the IEEE/CVF conference on computer vision and pattern recognition},
  pages={13693--13702},
  year={2021}
}

@article{tarvainen2017mean,
  title={Mean teachers are better role models: Weight-averaged consistency targets improve semi-supervised deep learning results},
  author={Tarvainen, Antti and Valpola, Harri},
  journal={Advances in neural information processing systems},
  volume={30},
  year={2017}
}

@article{laine2016temporal,
  title={Temporal ensembling for semi-supervised learning},
  author={Laine, Samuli and Aila, Timo},
  journal={arXiv preprint arXiv:1610.02242},
  year={2016}
}

@article{berthelot2019mixmatch,
  title={Mixmatch: A holistic approach to semi-supervised learning},
  author={Berthelot, David and Carlini, Nicholas and Goodfellow, Ian and Papernot, Nicolas and Oliver, Avital and Raffel, Colin A},
  journal={Advances in neural information processing systems},
  volume={32},
  year={2019}
}

@inproceedings{dai2023semi,
  title={Semi-supervised deep regression with uncertainty consistency and variational model ensembling via bayesian neural networks},
  author={Dai, Weihang and Li, Xiaomeng and Cheng, Kwang-Ting},
  booktitle={Proceedings of the AAAI Conference on Artificial Intelligence},
  volume={37},
  number={6},
  pages={7304--7313},
  year={2023}
}

@article{huang2024rankup,
  title={RankUp: Boosting Semi-Supervised Regression with an Auxiliary Ranking Classifier},
  author={Huang, Pin-Yen and Fu, Szu-Wei and Tsao, Yu},
  journal={arXiv preprint arXiv:2410.22124},
  year={2024}
}

@inproceedings{blum1998combining,
  title={Combining labeled and unlabeled data with co-training},
  author={Blum, Avrim and Mitchell, Tom},
  booktitle={Proceedings of the eleventh annual conference on Computational learning theory},
  pages={92--100},
  year={1998}
}

@inproceedings{bickel2004multi,
  title={Multi-view clustering.},
  author={Bickel, Steffen and Scheffer, Tobias},
  booktitle={ICDM},
  volume={4},
  number={2004},
  pages={19--26},
  year={2004},
  organization={Citeseer}
}

@article{gharaee2024bioscan,
  title={Bioscan-5m: A multimodal dataset for insect biodiversity},
  author={Gharaee, Zahra and Lowe, Scott C and Gong, ZeMing and Millan Arias, Pablo and Pellegrino, Nicholas and Wang, Austin T and Haurum, Joakim Bruslund and Eyriay, Iuliia and Kari, Lila and Steinke, Dirk and others},
  journal={Advances in Neural Information Processing Systems},
  volume={37},
  pages={36285--36313},
  year={2024}
}

@book{linnaeus1789systema,
  title={Systema Naturae per regna tria naturae, secundum classes, ordines, genera, species; cum characteribus, differentiis, synonymis, locis},
  author={Linnaeus, Carolus},
  volume={1},
  year={1789},
  publisher={apud JB Delamolliere}
}

@article{lu2016statistical,
  title={Statistical and computational guarantees of lloyd's algorithm and its variants},
  author={Lu, Yu and Zhou, Harrison H},
  journal={arXiv preprint arXiv:1612.02099},
  year={2016}
}

@article{putrama2024heterogeneous,
  title={Heterogeneous data integration: Challenges and opportunities},
  author={Putrama, I Made and Martinek, P{\'e}ter},
  journal={Data in Brief},
  pages={110853},
  year={2024},
  publisher={Elsevier}
}

@article{wafa2024integration,
  title={Integration Approaches for Heterogeneous Big Data: A Survey},
  author={Wafa’Za’al Alma’aitah, Addy and Quraan, Fatima N and Alkhawaldeh, Rami S and Alazab, Moutaz and Awajan, Albara},
  journal={Cybernetics and Information Technologies},
  volume={24},
  number={1},
  year={2024}
}

@article{van2020survey,
  title={A survey on semi-supervised learning},
  author={Van Engelen, Jesper E and Hoos, Holger H},
  journal={Machine learning},
  volume={109},
  number={2},
  pages={373--440},
  year={2020},
  publisher={Springer}
}

@article{yang2022survey,
  title={A survey on deep semi-supervised learning},
  author={Yang, Xiangli and Song, Zixing and King, Irwin and Xu, Zenglin},
  journal={IEEE transactions on knowledge and data engineering},
  volume={35},
  number={9},
  pages={8934--8954},
  year={2022},
  publisher={IEEE}
}

@incollection{ratnasingham2024bold,
  title={BOLD v4: A centralized bioinformatics platform for DNA-based biodiversity data},
  author={Ratnasingham, Sujeevan and Wei, Catherine and Chan, Dean and Agda, Jireh and Agda, Josh and Ballesteros-Mejia, Liliana and Boutou, Hamza Ait and El Bastami, Zak Mohammad and Ma, Eddie and Manjunath, Ramya and others},
  booktitle={DNA barcoding: Methods and protocols},
  pages={403--441},
  year={2024},
  publisher={Springer}
}

@article{sayers2019genbank,
  title={GenBank},
  author={Sayers, Eric W and Cavanaugh, Mark and Clark, Karen and Ostell, James and Pruitt, Kim D and Karsch-Mizrachi, Ilene},
  journal={Nucleic acids research},
  volume={47},
  number={D1},
  pages={D94--D99},
  year={2019},
}

@article{sohn2020fixmatch,
  title={Fixmatch: Simplifying semi-supervised learning with consistency and confidence},
  author={Sohn, Kihyuk and Berthelot, David and Carlini, Nicholas and Zhang, Zizhao and Zhang, Han and Raffel, Colin A and Cubuk, Ekin Dogus and Kurakin, Alexey and Li, Chun-Liang},
  journal={Advances in neural information processing systems},
  volume={33},
  pages={596--608},
  year={2020}
}

@article{wetzel2022twin,
  title={Twin neural network regression is a semi-supervised regression algorithm},
  author={Wetzel, Sebastian J and Melko, Roger G and Tamblyn, Isaac},
  journal={Machine Learning: Science and Technology},
  volume={3},
  number={4},
  pages={045007},
  year={2022},
  publisher={IOP Publishing}
}

@article{zhou2003learning,
  title={Learning with local and global consistency},
  author={Zhou, Dengyong and Bousquet, Olivier and Lal, Thomas and Weston, Jason and Sch{\"o}lkopf, Bernhard},
  journal={Advances in neural information processing systems},
  volume={16},
  year={2003}
}

@article{dasgupta2001pac,
  title={PAC generalization bounds for co-training},
  author={Dasgupta, Sanjoy and Littman, Michael and McAllester, David},
  journal={Advances in neural information processing systems},
  volume={14},
  year={2001}
}

@inproceedings{BalcanBlumYang2004,
  author    = {Maria-Florina Balcan and Avrim Blum and Ke Yang},
  title     = {Co-Training and Expansion: Towards Bridging Theory and Practice},
  booktitle = {Advances in Neural Information Processing Systems (NIPS)},
  year      = {2004}
}

@article{balcan2010discriminative,
  title={A discriminative model for semi-supervised learning},
  author={Balcan, Maria-Florina and Blum, Avrim},
  journal={Journal of the ACM (JACM)},
  volume={57},
  number={3},
  pages={1--46},
  year={2010},
  publisher={ACM New York, NY, USA}
}

@article{ReyRoth2012,
  author  = {Melanie Rey and Volker Roth},
  title   = {Copula Mixture Model for Dependency-Seeking Clustering},
  journal = {arXiv preprint arXiv:1206.6433},
  year    = {2012}
}

@inproceedings{ViinikanojaKlamiKaski2010,
  title={Variational Bayesian mixture of robust CCA models},
  author={Viinikanoja, Jaakko and Klami, Arto and Kaski, Samuel},
  booktitle={Joint European Conference on Machine Learning and Knowledge Discovery in Databases},
  pages={370--385},
  year={2010},
  organization={Springer}
}

@article{LeiMillerPinskyDubrawski2017,
  title   = {Characterization of Hemodynamic Signal by Learning Multi-View Relationships},
  author  = {Lei, Eric C. and Miller, Kyle and Pinsky, Michael R. and Dubrawski, Artur},
  journal = {arXiv preprint arXiv:1709.05602},
  year    = {2017}
}

@article{AzadkiaBalabdaoui2024,
  author  = {Mona Azadkia and Fadoua Balabdaoui},
  title   = {Linear Regression With Unmatched Data: A Deconvolution Perspective},
  journal = {Journal of Machine Learning Research},
  volume  = {25},
  number  = {197},
  pages   = {1--55},
  year    = {2024}
}

@article{RigolletWeed2019,
  author  = {Philippe Rigollet and Jonathan Weed},
  title   = {Uncoupled Isotonic Regression via Minimum Wasserstein Deconvolution},
  journal = {Information and Inference: A Journal of the IMA},
  volume  = {8},
  number  = {4},
  pages   = {691--717},
  year    = {2019},
  doi     = {10.1093/imaiai/iaz006}
}

@inproceedings{CarpentierSchluter2016,
  author    = {Alexandra Carpentier and Teresa Schl\"uter},
  title     = {Learning Relationships Between Data Obtained Independently},
  booktitle = {Proceedings of the 19th International Conference on Artificial Intelligence and Statistics (AISTATS)},
  year      = {2016}
}

@article{LampleConneauDenoyerRanzato2017,
  author  = {Guillaume Lample and Alexis Conneau and Ludovic Denoyer and Marc'Aurelio Ranzato},
  title   = {Unsupervised Machine Translation Using Monolingual Corpora Only},
  journal = {arXiv preprint arXiv:1711.00043},
  year    = {2017}
}

@inproceedings{ZhuParkIsolaEfros2017,
  author    = {Jun-Yan Zhu and Taesung Park and Phillip Isola and Alexei A. Efros},
  title     = {Unpaired Image-to-Image Translation using Cycle-Consistent Adversarial Networks},
  booktitle = {International Conference on Computer Vision (ICCV)},
  year      = {2017}
}

@inproceedings{HuangLiuBelongieKautz2018,
  author    = {Xun Huang and Ming-Yu Liu and Serge Belongie and Jan Kautz},
  title     = {Multimodal Unsupervised Image-to-Image Translation},
  booktitle = {European Conference on Computer Vision (ECCV)},
  pages     = {172--189},
  year      = {2018}
}

@article{Dempster1977,
  title={Maximum likelihood from incomplete data via the EM algorithm},
  author={Dempster, Arthur P and Laird, Nan M and Rubin, Donald B},
  journal={Journal of the royal statistical society: series B (methodological)},
  volume={39},
  number={1},
  pages={1--22},
  year={1977},
  publisher={Wiley Online Library}
}

@article{reimers2019sentence,
  title={Sentence-bert: Sentence embeddings using siamese bert-networks},
  author={Reimers, Nils and Gurevych, Iryna},
  journal={arXiv preprint arXiv:1908.10084},
  year={2019}
}

@article{young2014image,
  title={From image descriptions to visual denotations: New similarity metrics for semantic inference over event descriptions},
  author={Young, Peter and Lai, Alice and Hodosh, Micah and Hockenmaier, Julia},
  journal={Transactions of the association for computational linguistics},
  volume={2},
  pages={67--78},
  year={2014},
  publisher={MIT Press One Rogers Street, Cambridge, MA 02142-1209, USA journals-info~…}
}

@article{schubert2017dbscan,
  title={DBSCAN revisited, revisited: why and how you should (still) use DBSCAN},
  author={Schubert, Erich and Sander, J{\"o}rg and Ester, Martin and Kriegel, Hans Peter and Xu, Xiaowei},
  journal={ACM Transactions on Database Systems (TODS)},
  volume={42},
  number={3},
  pages={1--21},
  year={2017},
  publisher={Acm New York, NY, USA}
}

@inproceedings{srinivasan2021wit,
  title={Wit: Wikipedia-based image text dataset for multimodal multilingual machine learning},
  author={Srinivasan, Krishna and Raman, Karthik and Chen, Jiecao and Bendersky, Michael and Najork, Marc},
  booktitle={Proceedings of the 44th international ACM SIGIR conference on research and development in information retrieval},
  pages={2443--2449},
  year={2021}
}

@inproceedings{lin2014microsoft,
  title={Microsoft coco: Common objects in context},
  author={Lin, Tsung-Yi and Maire, Michael and Belongie, Serge and Hays, James and Perona, Pietro and Ramanan, Deva and Doll{\'a}r, Piotr and Zitnick, C Lawrence},
  booktitle={European conference on computer vision},
  pages={740--755},
  year={2014},
  organization={Springer}
}

@inproceedings{tan2019efficientnet,
  title={Efficientnet: Rethinking model scaling for convolutional neural networks},
  author={Tan, Mingxing and Le, Quoc},
  booktitle={International conference on machine learning},
  pages={6105--6114},
  year={2019},
  organization={PMLR}
}

@article{arias2023barcodebert,
  title={{BarcodeBERT}: Transformers for Biodiversity Analysis},
  author={Pablo Millan Arias
    and Niousha Sadjadi
    and Monireh Safari
    and ZeMing Gong
    and Austin T. Wang
    and Joakim Bruslund Haurum
    and Iuliia Zarubiieva
    and Dirk Steinke
    and Lila Kari
    and Angel X. Chang
    and Scott C. Lowe
    and Graham W. Taylor
  },
  journal={arXiv preprint arXiv:2311.02401},
  year={2023},
  eprint={2311.02401},
  archivePrefix={arXiv},
  primaryClass={cs.LG},
  doi={10.48550/arxiv.2311.02401},
}

@inproceedings{he2016deep,
  title={Deep residual learning for image recognition},
  author={He, Kaiming and Zhang, Xiangyu and Ren, Shaoqing and Sun, Jian},
  booktitle={Proceedings of the IEEE conference on computer vision and pattern recognition},
  pages={770--778},
  year={2016}
}

@article{mulcahy2022specimen,
  title={Specimen Identifiers: Linking tissues, DNA samples, and sequence data to voucher specimens in publicly accessible databases},
  author={Mulcahy, Daniel},
  journal={Biodiversity Information Science and Standards},
  volume={6},
  pages={e94625},
  year={2022},
  publisher={Pensoft Publishers}
}

@article{memoli2011gromov,
  title={Gromov--Wasserstein distances and the metric approach to object matching},
  author={M{\'e}moli, Facundo},
  journal={Foundations of computational mathematics},
  volume={11},
  number={4},
  pages={417--487},
  year={2011},
  publisher={Springer}
}

@article{cuturi2013sinkhorn,
  title={Sinkhorn distances: Lightspeed computation of optimal transport},
  author={Cuturi, Marco},
  journal={Advances in neural information processing systems},
  volume={26},
  year={2013}
}

@article{peyre2019computational,
  title={Computational optimal transport: With applications to data science},
  author={Peyr{\'e}, Gabriel and Cuturi, Marco},
  journal={Foundations and Trends{\textregistered} in Machine Learning},
  volume={11},
  number={5-6},
  pages={355--607},
  year={2019}
}

@inproceedings{peyre2016gromov,
  title={Gromov-wasserstein averaging of kernel and distance matrices},
  author={Peyr{\'e}, Gabriel and Cuturi, Marco and Solomon, Justin},
  booktitle={International conference on machine learning},
  pages={2664--2672},
  year={2016},
  organization={PMLR}
}
\bibliographystyle{icml2026}

\newpage
\appendix
\onecolumn

\section{Experiment Details}
\label{app:details}
This appendix details the embedding models, clustering algorithms, and parameters used.\footnote{The runtime and memory requirements depend heavily on the chosen clustering algorithm; our current implementation is not optimized, which complicates empirical evaluation.}

\subsection{BIOSCAN-5M}

BIOSCAN-5M contains the specimens images, taxonomic labels, and raw nucleotide barcode sequences of over 5 million insect samples \citep{gharaee2024bioscan}. For experimentation, our models will be given the encoded specimens images ($x$) to predict the encoded nucleotide barcodes ($y$), based on the latent taxonomic labels ($t$). Every sample is labeled according to the hierarchical Linnaean taxonomy: phylum, class, order, family, genus, and species. \citep{linnaeus1789systema}. 

We set up our dataset based on the test split of the original BIOSCAN-5M dataset, which is representative of the entire dataset and sizable (39,373) for our purposes. According to the lowest-level taxonomy label of "species", we can divide the dataset into hundreds of homogeneous species groups. Among those species, we randomly choose $C$ groups for each experiment run, prioritizing sampling species with different "family" labels, and then randomly draw 200 samples from each of the selected species groups. By selecting our species groups in this way, we approximately meet our theoretical assumption of sub-Gassian mixtures.

Within each group, samples are randomly assigned according to some fixed ratio to one of the three data-splits: input-only, output-only, and supervised.  For consistency, we set the output-only split to be 10\%, and vary the supervised percentage, using the rest as input-only data (When testing the inverse direction, we keep the input-only split to be 10\%, vary the supervised percentage, and use the rest as output-only data). Pooling all data, we remove the species information from the samples, hoping that our clustering procedures can group our data roughly according to latent species affiliations.

To convert the raw data into cluster-able data that we can easily train and evaluate on, we encode the specimens images with a EfficientNetB0 image encoder \citep{tan2019efficientnet} and the DNA barcodes with BarcodeBERT \citep{arias2023barcodebert}. We also narrow our sampling to the Insecta Class since it contains 98\% of the BIOSCAN-5M data.

\subsection{WIT}
Wikipedia Image-Text (WIT) \citep{srinivasan2021wit} Dataset is a large curated set of more than 37 million image-text associations extracted from Wikipedia articles. Each entry of the dataset contains an Wikipedia image with its image caption, as well as information of all pages containing this particular image.

To curate our dataset, we select the first 100,000 samples from WIT, encoding the images with ResNet50 \citep{he2016deep}and the text captions with Sentence-BERT\citep{reimers2019sentence}.

We further extract the English page titles associated with every image and encode them with Sentence-BERT. We use DBSCAN\citep{schubert2017dbscan} to discover dense clusters of datapoints with neighboring page title encodings, which suggests that these datapoints are associated with a latent category of Wikipedia pages. To further ensure closeness, we add another eligibility requirement for the clusters: every member in the cluster must share one word in their page title. Among these dense clusters retrieved, we randomly sample $C$ clusters, choose 25 samples from each cluster to be included in each run, and split them into three data-splits as in BIOSCAN (input-only split containing only images, output-only split containing only image captions, and the fully supervised split). This time we adjust the output-only percentage to 20\%, as the cluster sizes become small and it'll be challenging to form effective output clusters with only 10\% data. Pooling all data, we remove the page title information from the samples, hoping that our clustering procedures can group our data roughly according to the latent Wikipedia page categories.

\subsection{Flickr30k}

The Flickr30k dataset \citep{young2014image} is comprised of 31,783 images that capture people engaged in everyday activities and events. For each image, the dataset provides 5 human-annotated descriptive captions.

We select the first 10,000 samples from Flickr30k, and split each sample into 5 datapoints, so that each entry contains one unique caption and one (possibly repeated) image.

In close parallel to the WIT experiment, we encode the images with ResNet50 and the text captions with Sentence-BERT. To select meaningful families of data, we further encode the text captions with TF-IDF and perform clustering, so that caption that share key words would be clustered close together. We then use DBSCAN to discover dense clusters of datapoints, randomly sample $C$ clusters and sample 25 datapoints per cluster to be included each experiment run. We split the data in each cluster into three data-splits as before, then pool all data for the experiment.

A slight difference from WIT is in caption prediction, where we measure the distance between our model-predicted caption and the closest of the 5 captions that image corresponds to.

\subsection{COCO}

The Microsoft Common Objects in Context Dataset (COCO \citep{lin2014microsoft}) is another image-text dataset. Every image in the dataset is paired with an array of text annotations, as well as a list of categorical labels, corresponding to different objects present in the image.

Our curated dataset is extracted from the validation split of the original COCO dataset. We sample all datapoints with only one categorical label in their list, and group them according to their categorical labels. In each experiment run, we randomly sample C groups and sample 200 datapoints per group, then follow the experimental procedure as exactly described in Flickr30k.

\subsection{Note on baselines}
We adapted the baseline methods to our experiment environment, keeping faithful to the original methods and ensuring conceptual correctness. We allocated a reasonable fine-tuning budget for our baseline methods with a small grid across key hyper-parameters. Since \bc does not require any finetuning, we didn't spend excessive resource in the fine-tuning of the baselines, so the baselines used in our experiments may not be the optimized version. For baseline methods that involve iterations, we set a reasonable and generous maximum iteration according to the runtime of each algorithm, which is normally 2000, as compared to 100 or less maximum iterations for \bc. Please see our code for additional details. We also attach the tuning grid for reference.

\begin{table*}[t]
\centering
\small
\caption{Hyperparameter grids used for baseline fine-tuning. Each baseline has a small grid over key hyperparameters; \bc\ and KNN require no tuning. Iterative methods use a generous cap (typically $2000$ iterations) versus $\leq 100$ for \bc.}
\label{tab:hparam-grids}
\renewcommand{\arraystretch}{1.15}
\begin{tabularx}{\textwidth}{lX}
\toprule
\textbf{Method} & \textbf{Grid} \\
\midrule
FixMatch &
\ttfamily lr=\{1e-4, 3e-4, 1e-3\};\,
batch\_size=\{32, 64\};\,
alpha\_ema=\{0.99, 0.999\};\,
lambda\_u\_max=\{0.5, 1.0\};\,
rampup\_length=\{10, 30\};\,
conf\_threshold=\{0.05, 0.1\} \\
LapRLS &
\ttfamily lam=\{1e-5, 1e-3, 1e-1\};\,
gamma=\{1e-3, 1e-1, 1\};\,
k=\{5, 10, 20\};\,
sigma=\{0.5, 1.0, 2.0\} \\
TNNR &
\ttfamily rep\_dim=\{32, 64, 128\};\,
beta=\{0.01, 0.1, 1.0\};\,
lr=\{1e-4, 3e-4, 1e-3\} \\
TSVR &
\ttfamily C=\{0.1, 1, 10\};\,
epsilon=\{0.01, 0.1\};\,
gamma=\{0.1, 1\};\,
self\_training\_frac=\{0.1, 0.2, 0.5\} \\
UCVME &
\ttfamily lr=\{1e-4, 3e-4, 1e-3\};\,
w\_unl=\{1, 5, 10\};\,
mc\_T=\{5, 10\} \\
GCN &
\ttfamily hidden=\{32, 64, 128\};\,
dropout=\{0.0, 0.1, 0.3\};\,
lr=\{1e-3, 3e-3\} \\
KMM &
\ttfamily alpha=\{1e-2, 1e-1\};\,
kmm\_B=\{100, 1000\};\,
kmm\_eps=\{1e-3, 1e-2\};\,
sigma=\{0.5, 1.0\} \\
EM &
\ttfamily n\_components=\{2, 3\};\,
max\_iter=\{100, 200\};\,
tol=\{1e-3, 1e-4\};\,
eps=\{1e-3, 1e-4\} \\
EOT &
\ttfamily eps=\{1e-3, 1e-2, 1e-1, 1, 10\};\,
ridge\_alpha=\{1e-2, 1e-3, 1e-4\};\,
tol=\{1e-5, 1e-7, 1e-9\} \\
GW &
\ttfamily max\_iter=\{200, 400, 800\};\,
tol=\{1e-5, 1e-7, 1e-9\} \\
\bottomrule
\end{tabularx}
\end{table*}

\section{On the Importance of Representation or \bc} 
\label{app:interp}
Following our algorithm, we independently cluster the input space (e.g., encoded images) and the output space (e.g., encoded DNA or text). In expectation, effective clustering shall retrieve latent class affiliations. For example, in the BIOSCAN experiment where the latent variable is "species", samples assigned to the same input cluster shall have the same species label, as with output cluster. We measure this clustering effectiveness using Adjusted Mutual Information (AMI) between cluster assignments and latent labels for both inputs and outputs.

Since our algorithm is model-agnostic, we are free to experiment with a wide range of clustering methods. As we know the number of latent variables included in every experiment, hence the ideal number of clusters, k-clustering methods are natural choices: k-means, spectral clustering, and gaussian mixture model. We also explored agglomerative clustering as a baseline. As we discovered over iterative testing, constrained k-means with balanced cluster sizes is our best performer as measured through AMI, possibly due to the fact that we curated our dataset with equal species-cluster sizes.

Another metric that's directly related to the algorithm's effectiveness is the accuracy of bridged association learning. After clustering, as each supervised point is associated with one input and one output cluster, we obtain a voted mapping between input and output clusters. We measure the accuracy of this mapping by comparing it to the ground-truth. Since every input/output cluster roughly corresponds to one distinct latent variable, with high probability every latent variable will dominate exactly one input cluster and one output cluster, allowing us to obtain a non-overlapping ground-truth mapping through latent-by-latent reconstruction. After trials with different voting mechanisms, including Margin-based voting and Hungarian Algorithm, the most accurate mechanism turns out to be simple majority-based voting.

\section{\bc Algorithm (Full version with Inverse Prediction)}

This is an alternative version of our algorithm that would be potentially useful in practice, using the bidirectional property of Bridged Clustering.

\begin{algorithm}
\caption{\bc Algorithm (Full version with Inverse Prediction)}
\label{alg:bridged_clustering_full}
\textbf{Input}: Input-only set $\mathcal{X} = \{ x_1, x_2, \ldots, x_{|\mathcal{X}|} \}$; Output-only set $\mathcal{Y} = \{ y_1, y_2, \ldots, y_{|\mathcal{Y}|} \}$; \\sparse supervised set $\mathcal{S} = \{ (x'_i, y'_i) \}_{i=1}^k$; Input Test-set $\mathcal{X_\text{test}} = \{ x_1, x_2, \ldots, x_{|\mathcal{X_\text{test}}|} \}$; Output Test-set $\mathcal{Y_\text{test}} = \{ y_1, y_2, \ldots, y_{|\mathcal{Y_\text{test}}|} \}$ (Note: $\mathcal{X_\text{test}} = \mathcal{X}$, $\mathcal{Y_\text{test}} = \mathcal{Y}$ in the \emph{transductive} setting) \\
\textbf{Output}: $\hat{\mathcal{Y}} = \{ \hat{y}_1, \hat{y}_2, \ldots, \hat{y}_{|\mathcal{X_\text{test}}|} \},
\hat{\mathcal{X}} = \{ \hat{x}_1, \hat{x}_2, \ldots, \hat{x}_{|\mathcal{Y_\text{test}}|} \}$
\begin{algorithmic}[1]
\STATE \textbf{Call} Algorithm \ref{alg:bridged_clustering}
\STATE Using the supervised set $\mathcal{S}$, learn the reversed mapping $\hat A_{y \to x}$ between clusters in $\mathcal{Y}$ and clusters in $\mathcal{X}$.
\FOR{each sample $y_i$ in $\mathcal{Y_\text{test}}$}
    \STATE Assign $y_i$ to a cluster $c_y = \mathcal{C}_\mathcal{Y_\text{test}}(y_i)$.
    \STATE Find corresponding cluster in $\mathcal{X}$: $c_x = \hat A_{y \to x}(c_y)$.
    \STATE Predict $\hat{x}_i$ as the centroid of cluster $c_x$ in $\mathcal{X}$.
\ENDFOR
\STATE \textbf{return} Predicted outputs $\hat{\mathcal{Y}}$ for given inputs, Predicted inputs $\hat{\mathcal{X}}$ for given outputs. 
\end{algorithmic}
\end{algorithm}

\section{Runtime and Memory Complexity}
\label{sec:runtime}

We break down the cost of \bc into clustering, bridge learning, and inference.
To remain solver-agnostic, let $f(n,C,d)$ denote the time our chosen algorithm takes to cluster $n$ points in $\R^d$ into $C$ clusters.
For \bc:
\[
\text{training} = f(n_X,C,d)+f(n_Y,C,d')+O(k),\] \[
\text{per-query inference} = O(Cd + d'),
\]
with memory $O(n_X d + n_Y d' + C(d{+}d') {+} C^2)$.\footnote{If only a mapping $[C]\!\to\![C]$ is stored, the bridge term is $O(C)$; $O(C^2)$ is a safe upper bound for storing a sparse $C{\times}C$ vote/mapping matrix.}
Instantiating $f$: Lloyd/$k$-means++ gives $O(n C d I)$ where $I$ is the number of iterations; mini-batch $k$-means $O(m C d I)$ with batch $m\!\ll\! n$; pruned $k$-means $O(n C' d I)$ with $C'\!\le\!C$ effective distance calls.
Thus, training is linear in $n_X,n_Y$ when $C\!\ll\! n$, and inference requires only $O(C)$ distance computations instead of dense cross-space objects.

Now, as we found in our experiments in Section \ref{sec:experiments}, the transport-based methods, EOT and GW, are the closest competitors to our method.
We compare our method asymptotically to the standard implementations of EOT and GW. This analysis is conducted in the transductive setting, though the asymptotic conclusions extend similarly to the inductive case. Let $S$ denote the number of Sinkhorn iterations in EOT, and $T_{\text{GW}}$ the number of outer GW iterations; we report only the dominant terms.

\begin{center}
\setlength{\tabcolsep}{6pt}
\renewcommand{\arraystretch}{1.1}
\begin{tabular}{lcc}
\toprule
\textbf{Method} & \textbf{Training time} & \textbf{Memory} \\
\midrule
BC\ (ours) 
& $f(n_X,C,d){+}f(n_Y,C,d'){+}O(k)$
& $O(n_X d + n_Y d' + C(d{+}d'){+}C^2)$ \\
EOT 
& $O(S\,n_X n_Y)$
& $O(n_X n_Y)$ \\
GW 
& $O(T_{\text{GW}} (n_X^2 n_Y + n_X n_Y^2))$
& $O(n_X^2 + n_Y^2 + n_X n_Y)$ \\
\bottomrule
\end{tabular}
\end{center}

\noindent Since both transport-based methods involve Sinkhorn iterations and dense couplings, they incur at least quadratic cost in both time and memory, which scales quickly with data increase. In contrast, \bc with a fast clustering routine (e.g., mini-batch $k$-means) achieves training time linear in $n_X,n_Y$ and memory linear in the raw data size plus a (sparse) $C{\times}C$ bridge, making it more efficient in both runtime and memory than transport-based baselines in large-pool regimes.

\begin{figure}[H]
    \centering
    \includegraphics[width=0.56\linewidth]{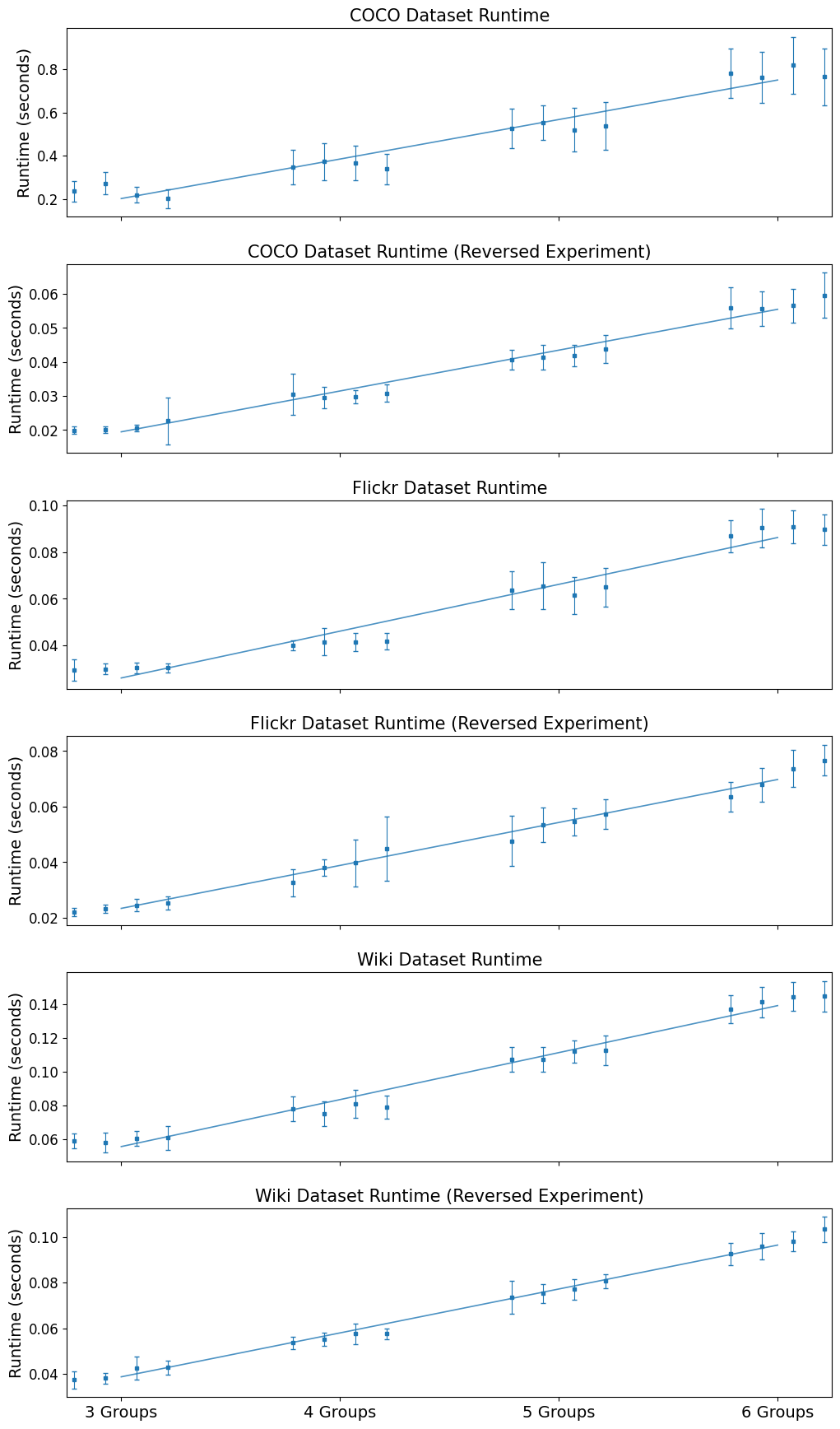}
    \caption{Empirical Runtime shows linear growth, mirroring theoretical bound (Inductive Experiments).}
    \label{fig:placeholder}
\end{figure}

\section{Detailed Experiment Results: MSE}
\label{app:mse}
\begin{figure*}[h]
    \centering
    \includegraphics[width=.65\linewidth]{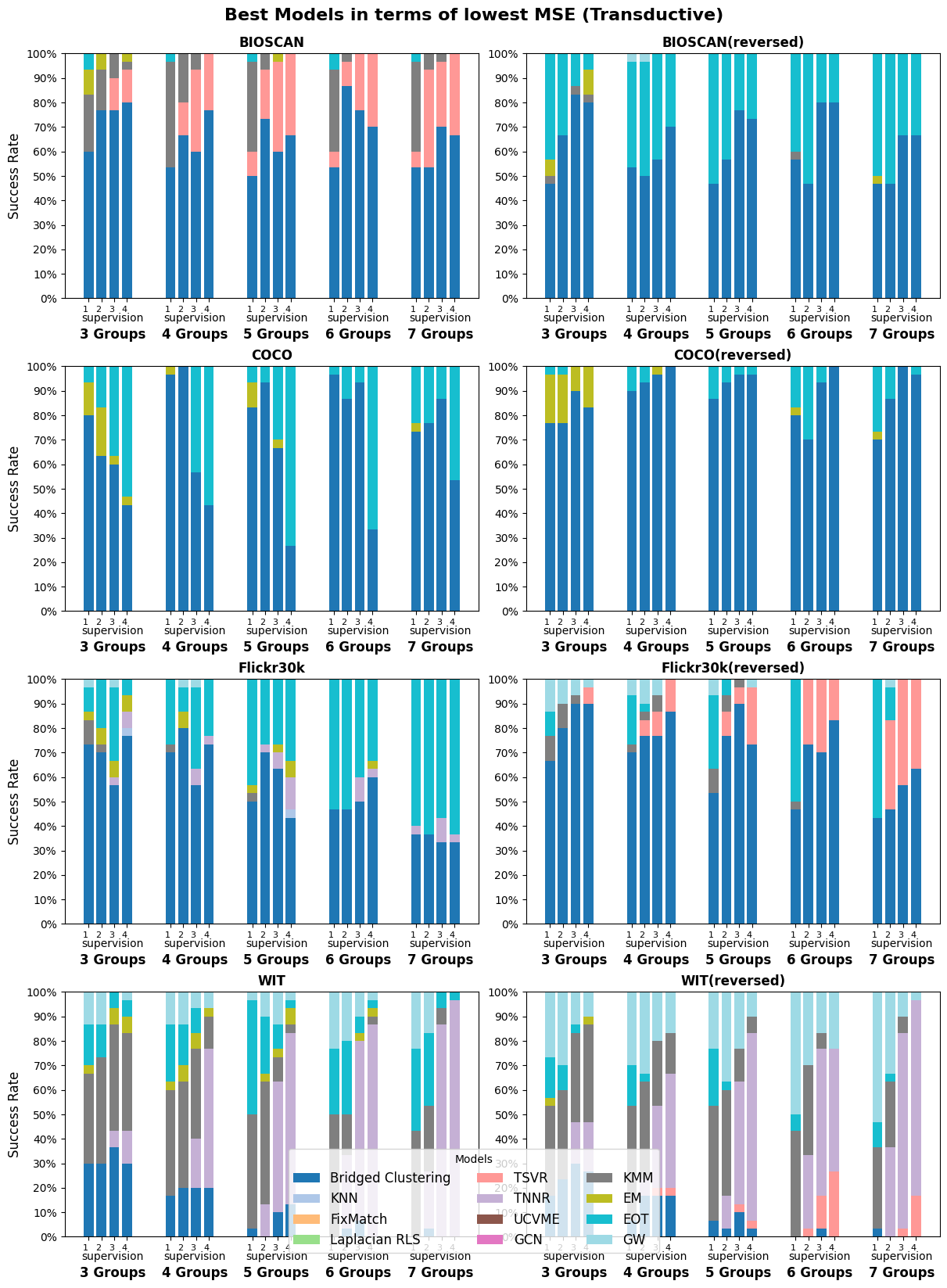}
    \caption{Tansductive Experiment: Best models in terms of lowest MSE, computed across 30 randomized trials per setting. Each bar represents the 30 trials of one setting. For example, if \bc achieves the lowest MSE among all models in 15 out of the 30 trials for some setting, the bar that corresponds to that setting will be colored 50\% blue. The 1,2,3,4 ticks on the bottom represent the settings with 1,2,3,4 supervised samples per cluster.}
    \label{fig:figa}
\end{figure*}
\begin{figure}[h]
    \centering
    \includegraphics[width=.75\linewidth]{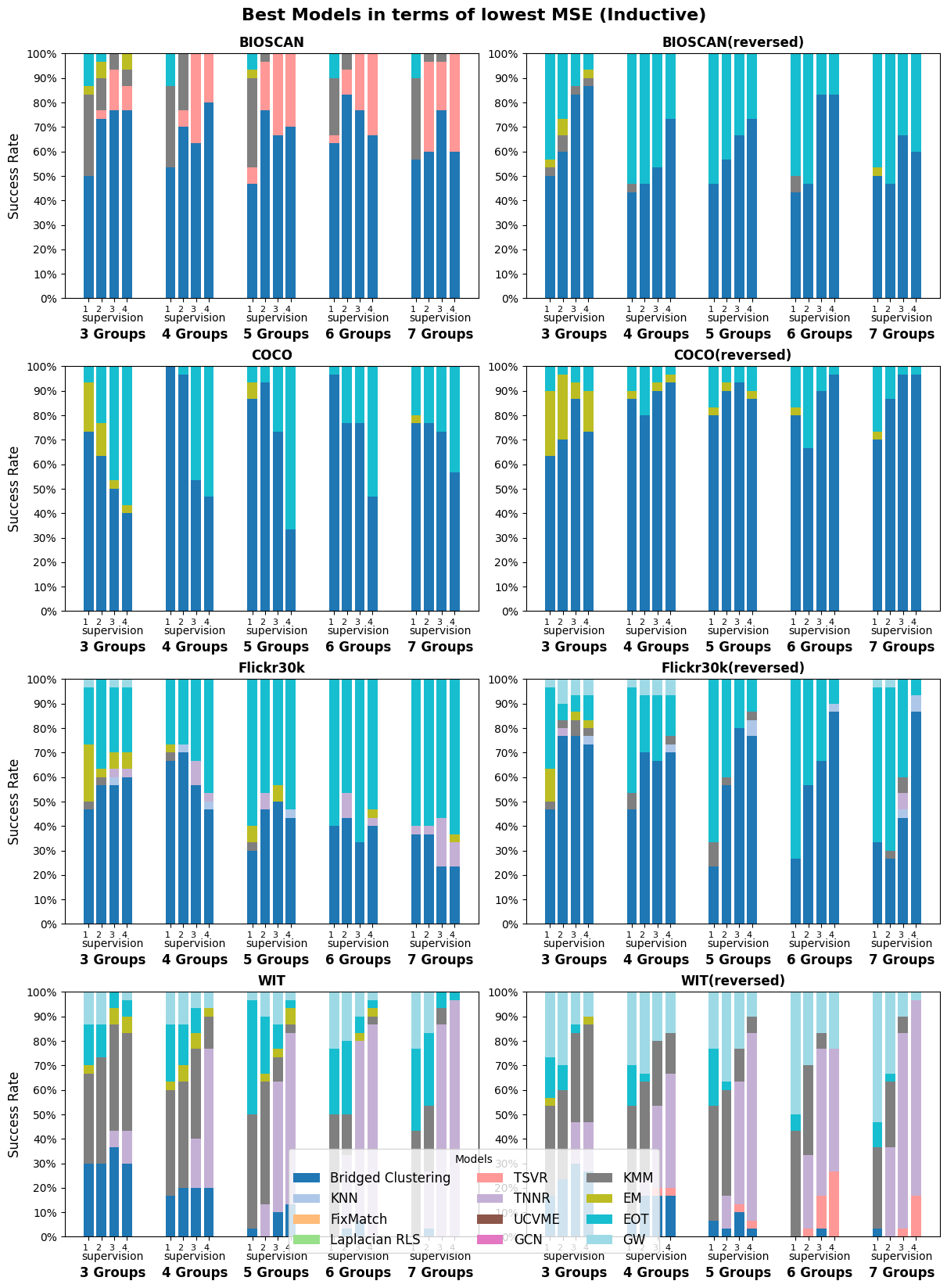}
    \caption{Inductive Experiment: Best models in terms of lowest MSE, computed across 30 randomized trials per setting. Each bar represents the 30 trials of one setting. For example, if \bc achieves the lowest MSE among all models in 15 out of the 30 trials for some setting, the bar that corresponds to that setting will be colored 50\% blue. The 1,2,3,4 ticks on the bottom represent the settings with 1,2,3,4 supervised samples per cluster.}
    \label{fig:figb}
\end{figure}
\begin{figure*}[h]
    \centering
    \includegraphics[width=0.75\linewidth]{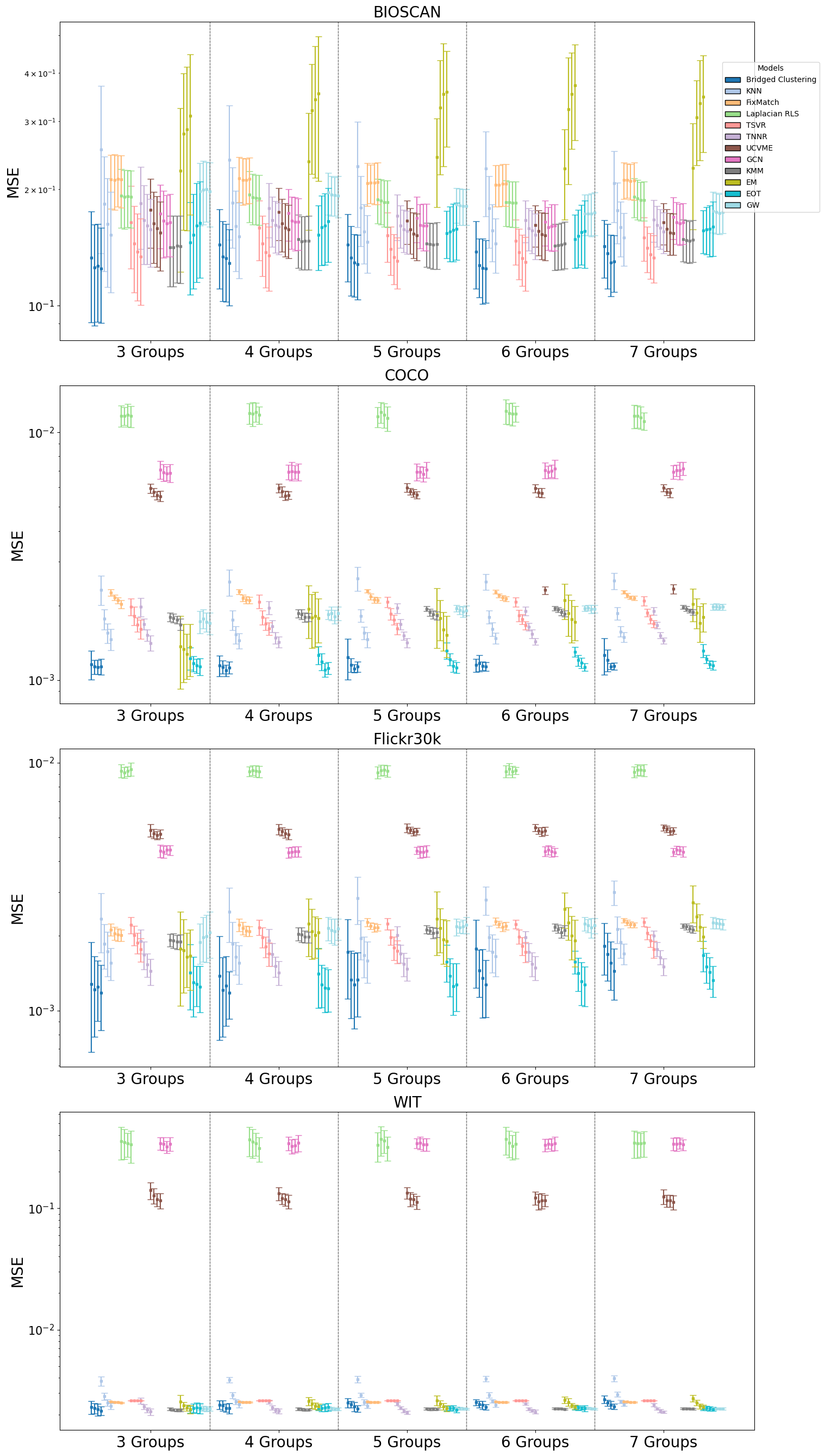}
    \caption{MSE distribution of different models in the transductive setting. The 4 distribution plots of the same color represent the settings with 1,2,3,4 supervised samples per cluster.}
    \label{fig:fig4}
\end{figure*}
\begin{figure*}[h]
    \centering
    \includegraphics[width=0.7\linewidth]{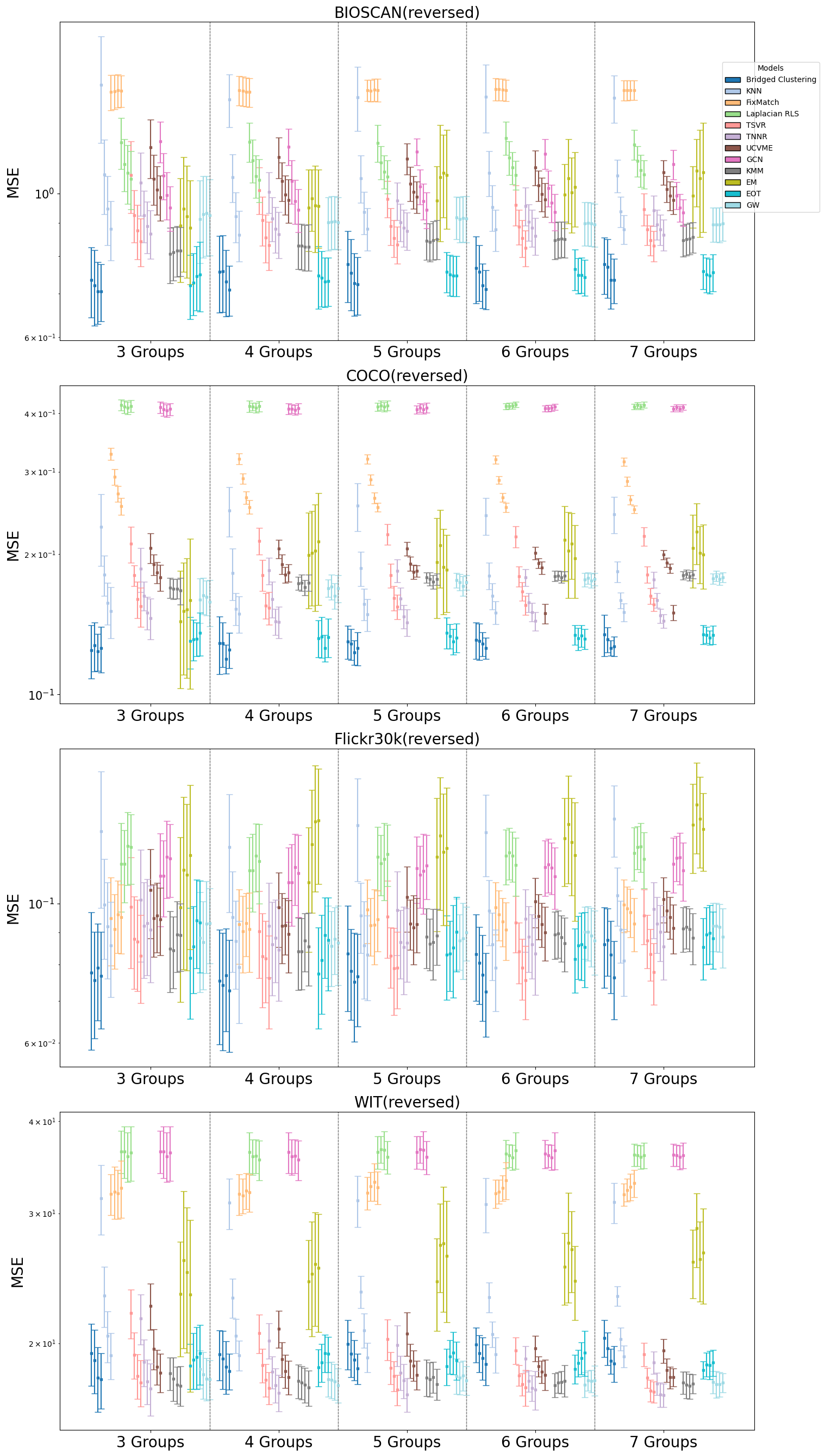}
    \caption{MSE distribution of different models in reversed experiments in the transductive setting. The 4 distribution plots of the same color represent the settings with 1,2,3,4 supervised samples per cluster.}
    \label{fig:fig5}
\end{figure*}
\begin{figure}
    \centering
    \includegraphics[width=0.75\linewidth]{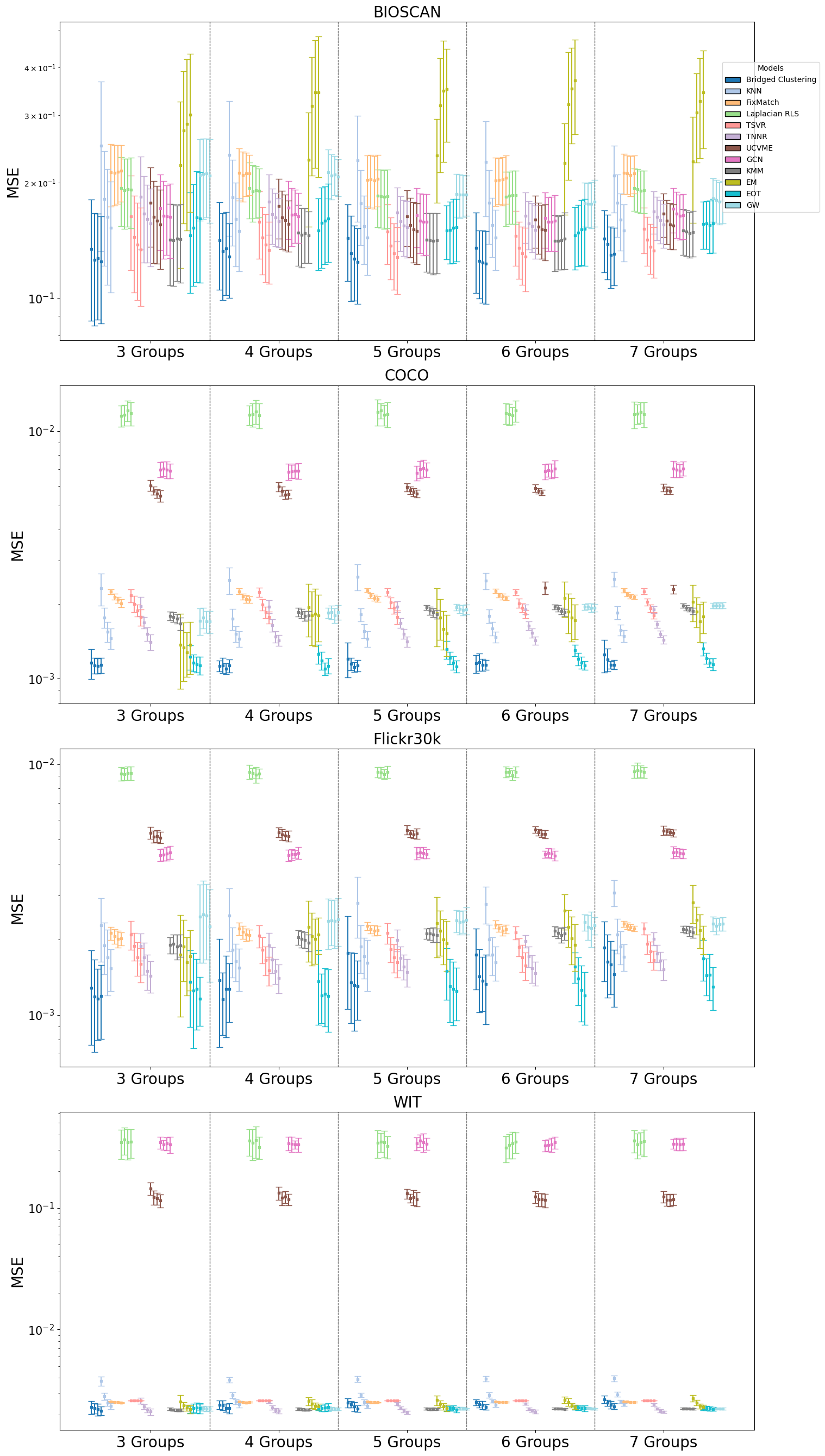}
    \caption{MSE distribution of different models in the inductive setting. The 4 distribution plots of the same color represent the settings with 1,2,3,4 supervised samples per cluster.}
    \label{fig:placeholder}
\end{figure}
\begin{figure}
    \centering
    \includegraphics[width=0.75\linewidth]{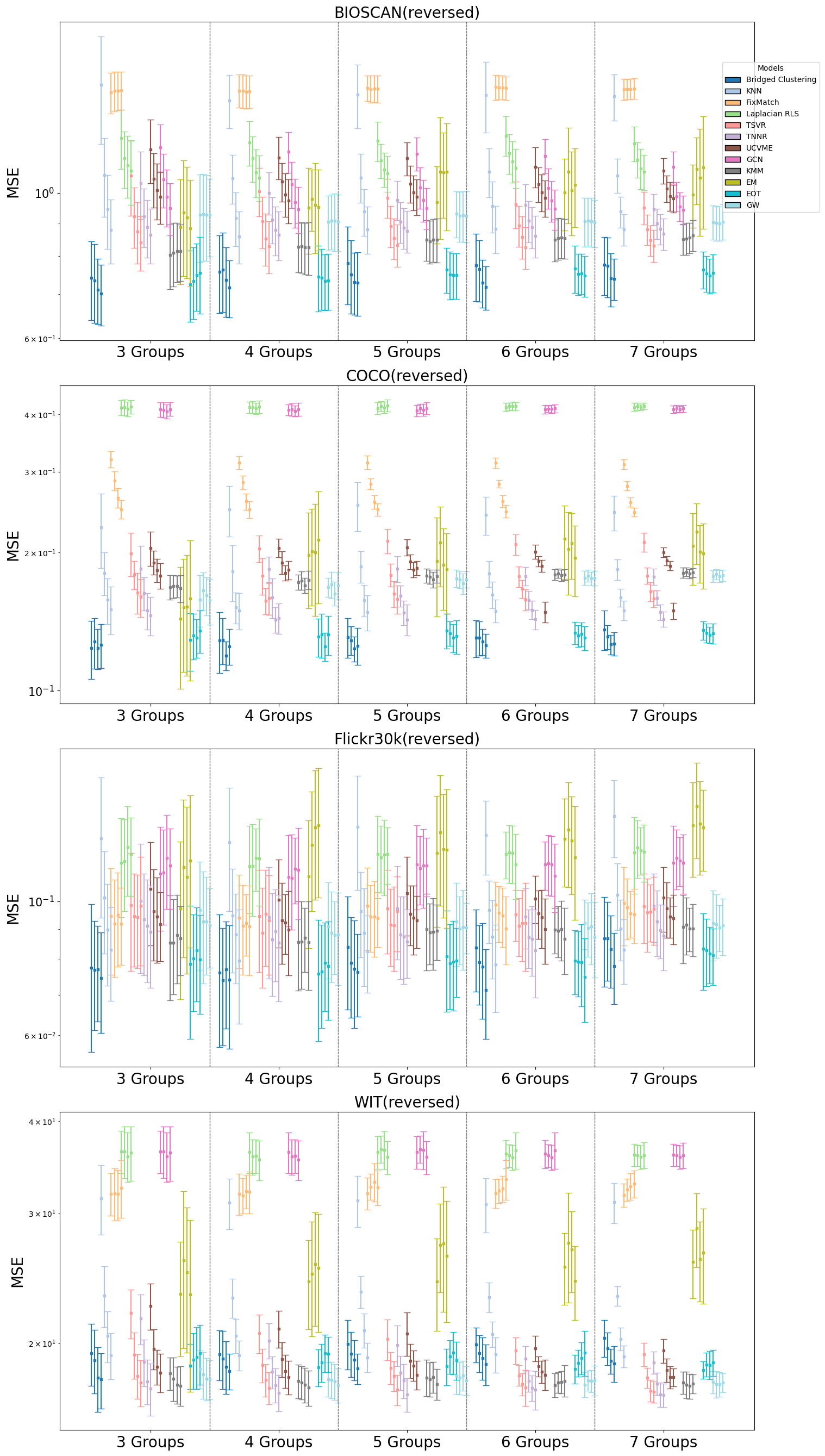}
    \caption{MSE distribution of different models in reversed experiments in the inductive setting. The 4 distribution plots of the same color represent the settings with 1,2,3,4 supervised samples per cluster.}
    \label{fig:figf}
\end{figure}

\clearpage
\section{Detailed Experiment Results: Cluster Quality and Accuracy (Transductive)}
\label{app:cluster_quality}

\begin{figure*}[h]
    \centering
    \includegraphics[width=0.88\linewidth]{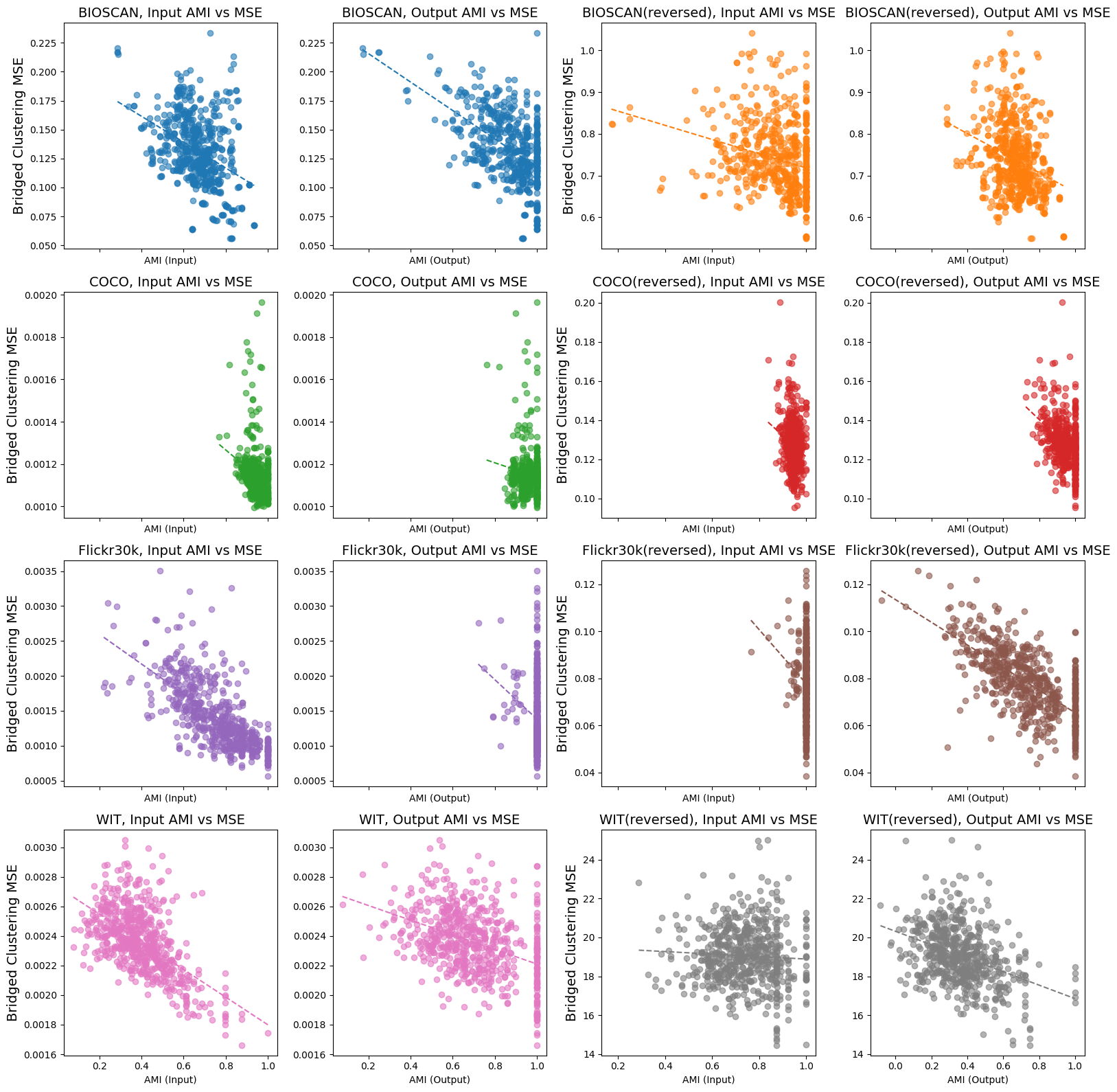}
    \caption{Mean Squared Error of \bc V.S. Adjusted Mutual Information of Clusters.}
    \label{fig:fig2}
\end{figure*}

\begin{figure*}
    \centering
    \includegraphics[width=0.7\linewidth]{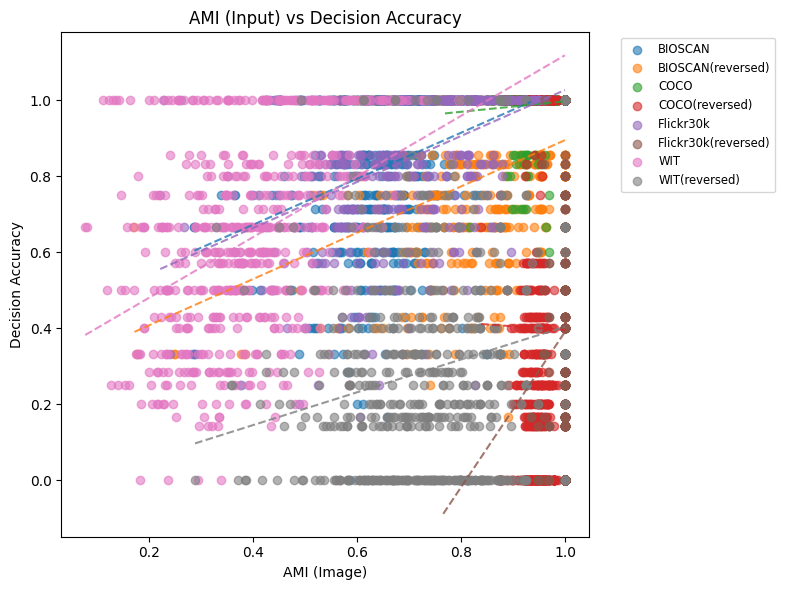}
    \caption{Bridging Accuracy as function of Adjusted Mutual Information of input clusters.}
    \label{fig:figh}
\end{figure*}

\begin{figure*}
    \centering
    \includegraphics[width=0.7\linewidth]{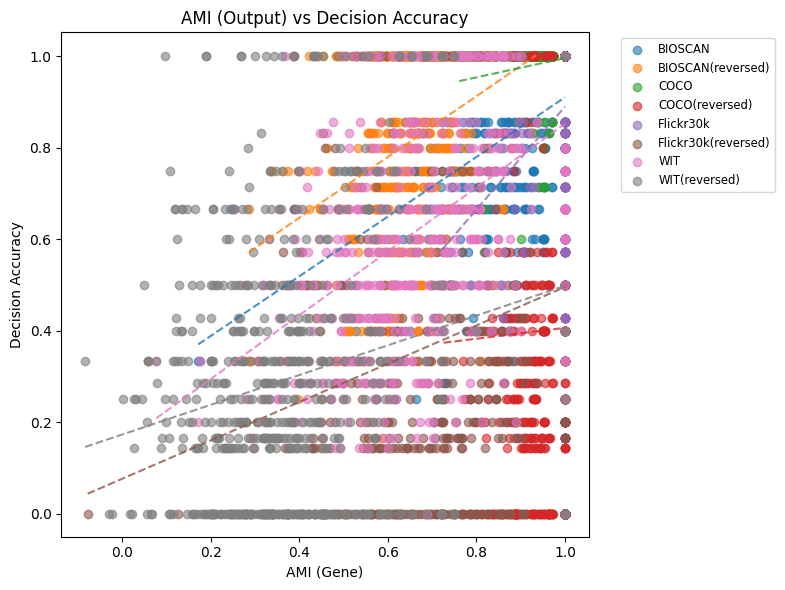}
    \caption{Bridging Accuracy as function of Adjusted Mutual Information of output clusters.}
    \label{fig:figi}
\end{figure*}

\begin{figure}
    \centering
\includegraphics[width=0.6\linewidth]{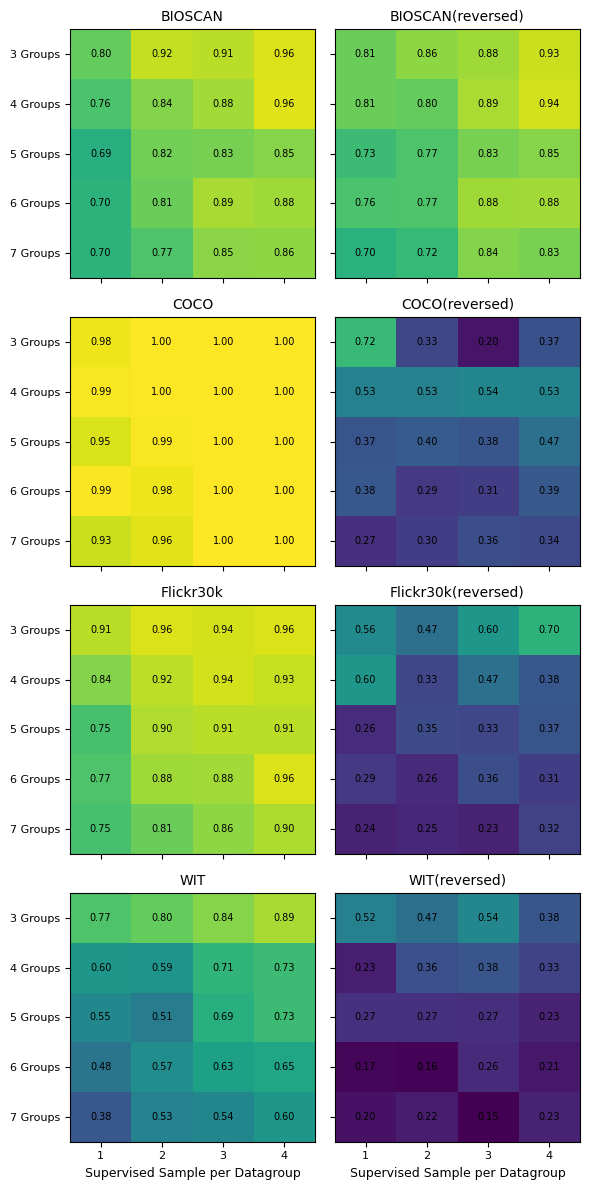}
    \caption{Average Bridging Accuracy.}
    \label{fig:bridg_acc}
\end{figure}

\section{Experiment Results for Algorithmic Variants}
\label{app:variant}
\begin{figure}[H]
    \centering
    \includegraphics[width=0.65\linewidth]{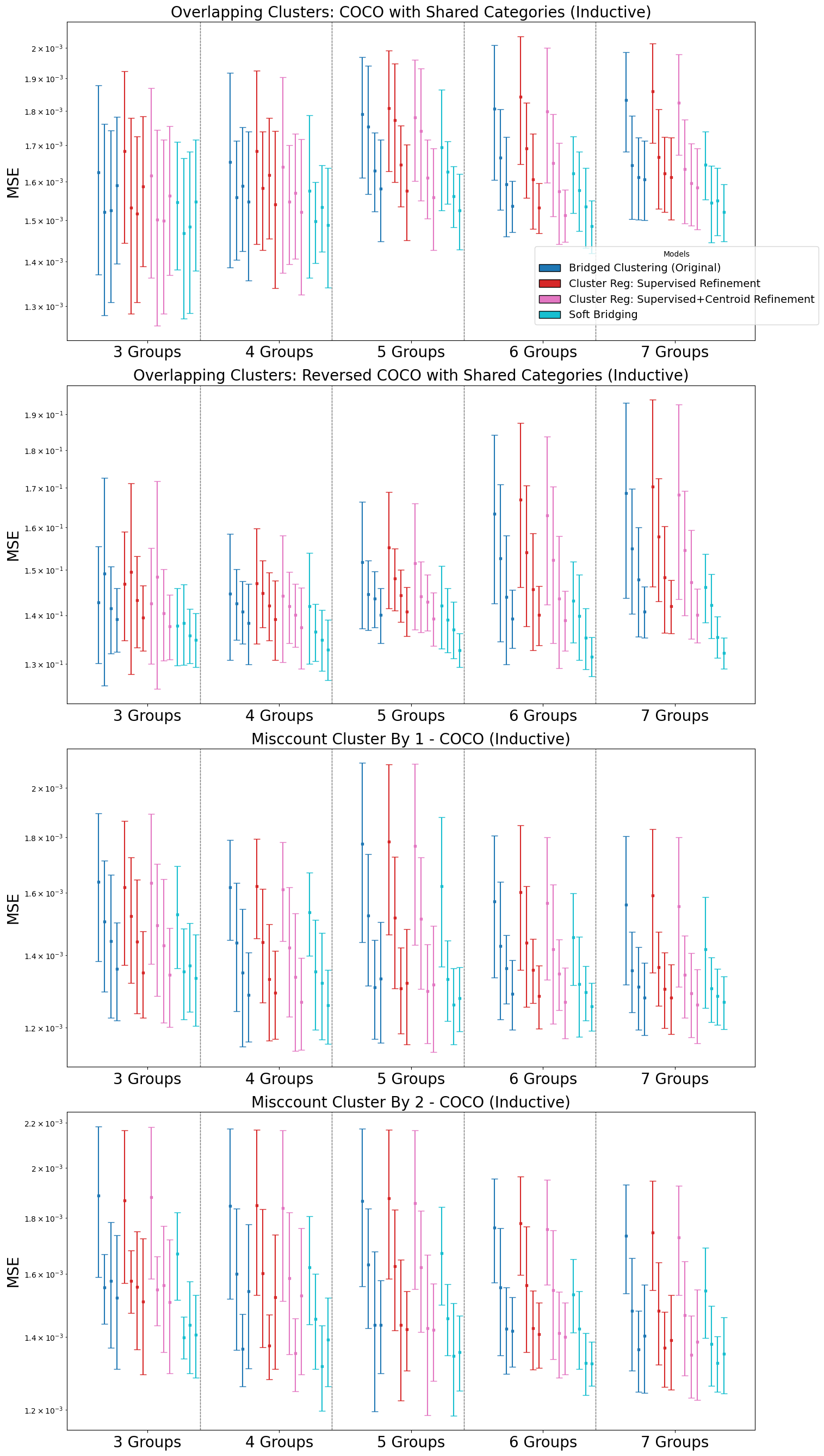}
        \caption{Inductive Experiments using Algorithmic Variants: The 4 distribution plots of the same color represent the settings with 1,2,3,4 supervised samples per cluster. Notice that Soft-Bridging is the overall strongest method in addressing both overlapping and misccounted clusters. For Cluster-wise Regression, Supervised Refinement alone performs unstably, but adding centroid to regressor fitting improves model performance, slightly outperforming the original Bridged Clustering model. Here, miscounting refers to overcounting the number of clusters, and we note that undercounting experiments show similar results.}
    \label{fig:figk}
\end{figure}

\begin{figure}[H]
    \centering
    \includegraphics[width=0.8\linewidth]{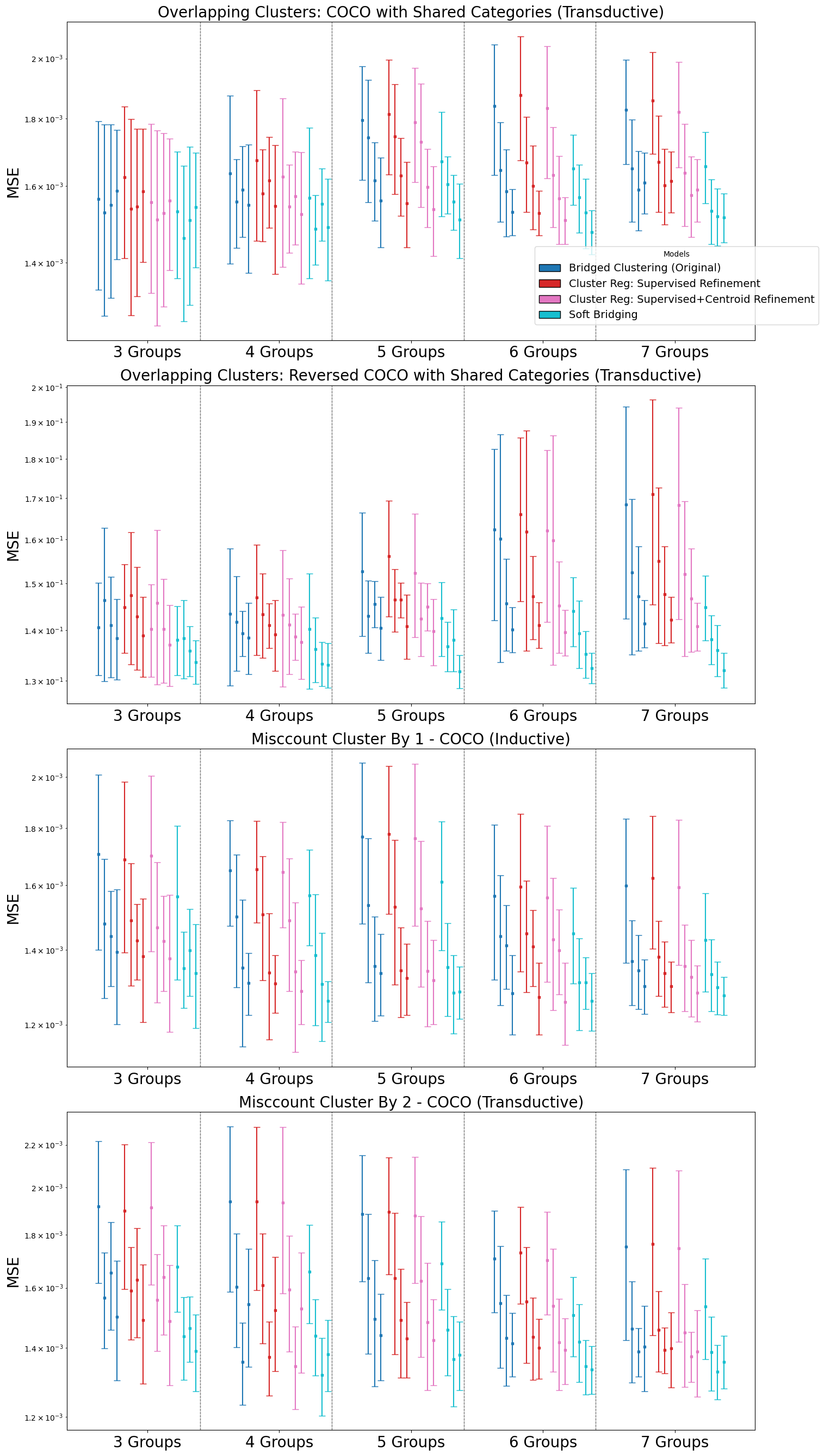}
    \caption{Transductive Experiments using Algorithmic Variants. Similar Observations with Inductive Setting.}
    \label{fig:figl}
\end{figure}

\section{Experiment Results for Adversarial Conditions}
\label{app:adver}
Our base model Bridged Clustering does incur more loss in cluster-adversarial conditions, but still maintains overall competitive performance, among the strongest models with low MSE.

\begin{figure}[H]
    \centering
    \includegraphics[width=0.63\linewidth]{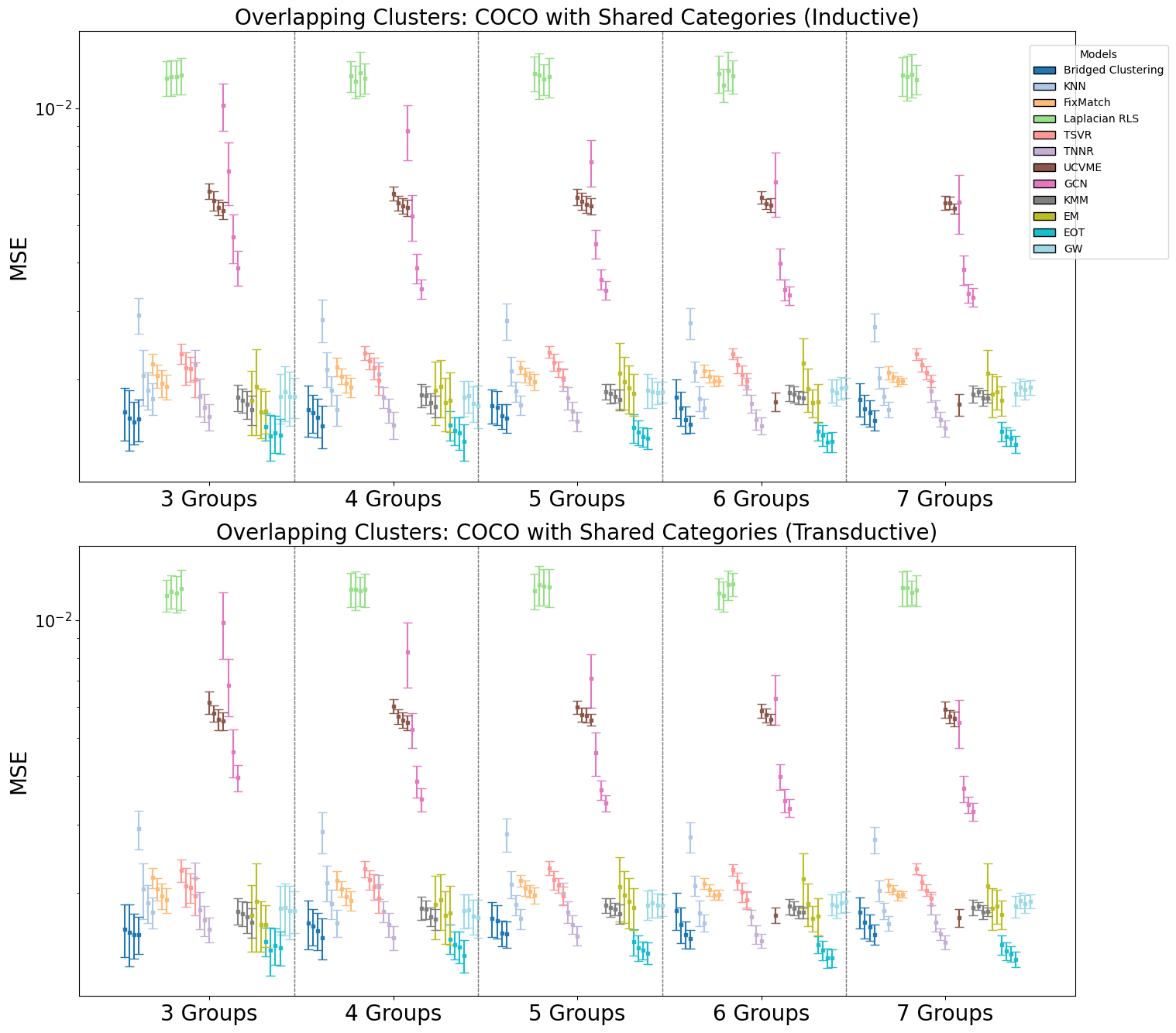}
    \caption{Running main experiment on COCO with Overlapping Data Classes}
    \label{fig:figm}
\end{figure}

\begin{figure}[H]
    \centering
    \includegraphics[width=0.63\linewidth]{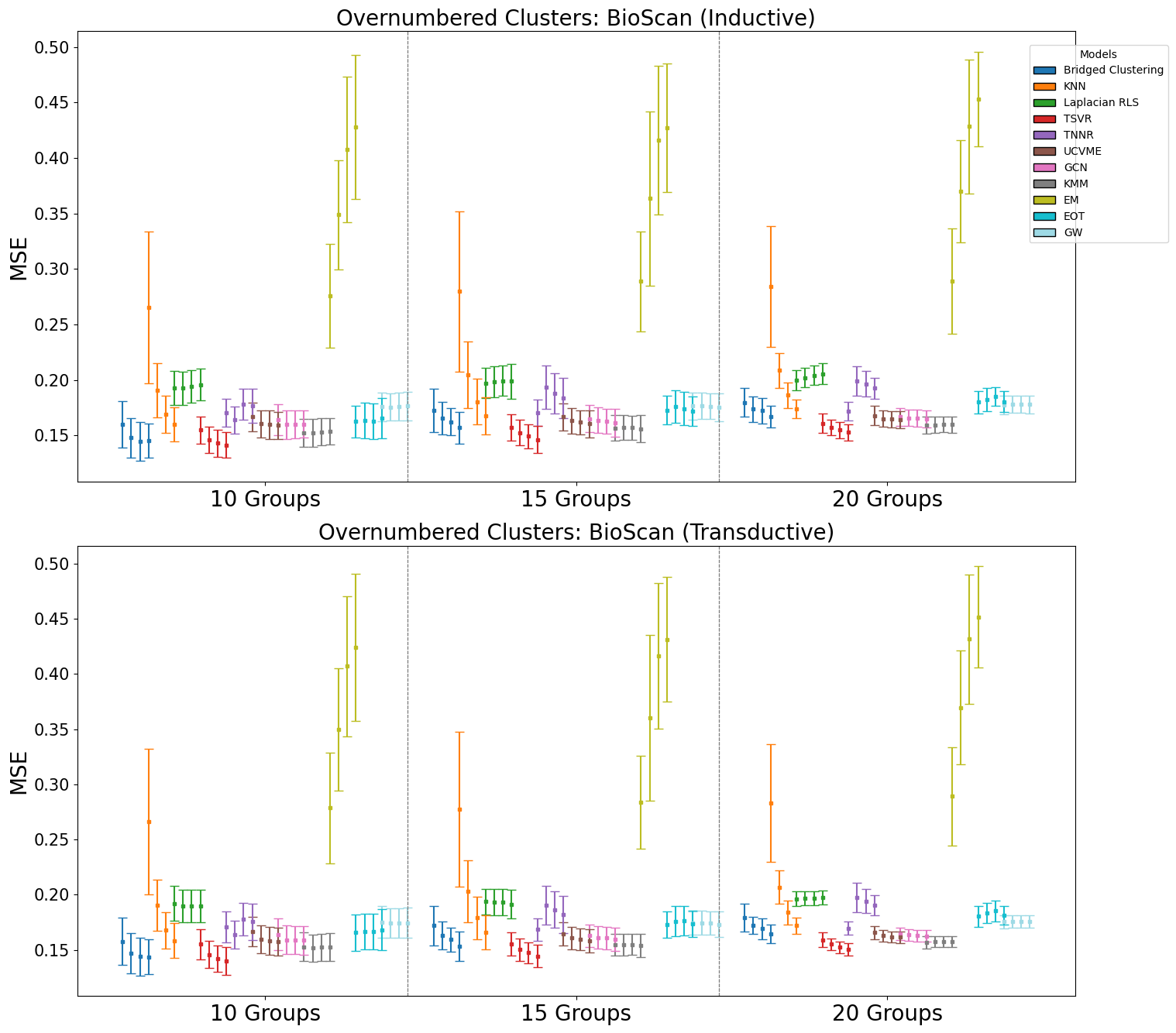}
        \caption{Running main experiment on Large Cluster Counts for BIOSCAN}
    \label{fig:fign}
\end{figure}

\begin{figure}
    \centering
    \includegraphics[width=0.77\linewidth]{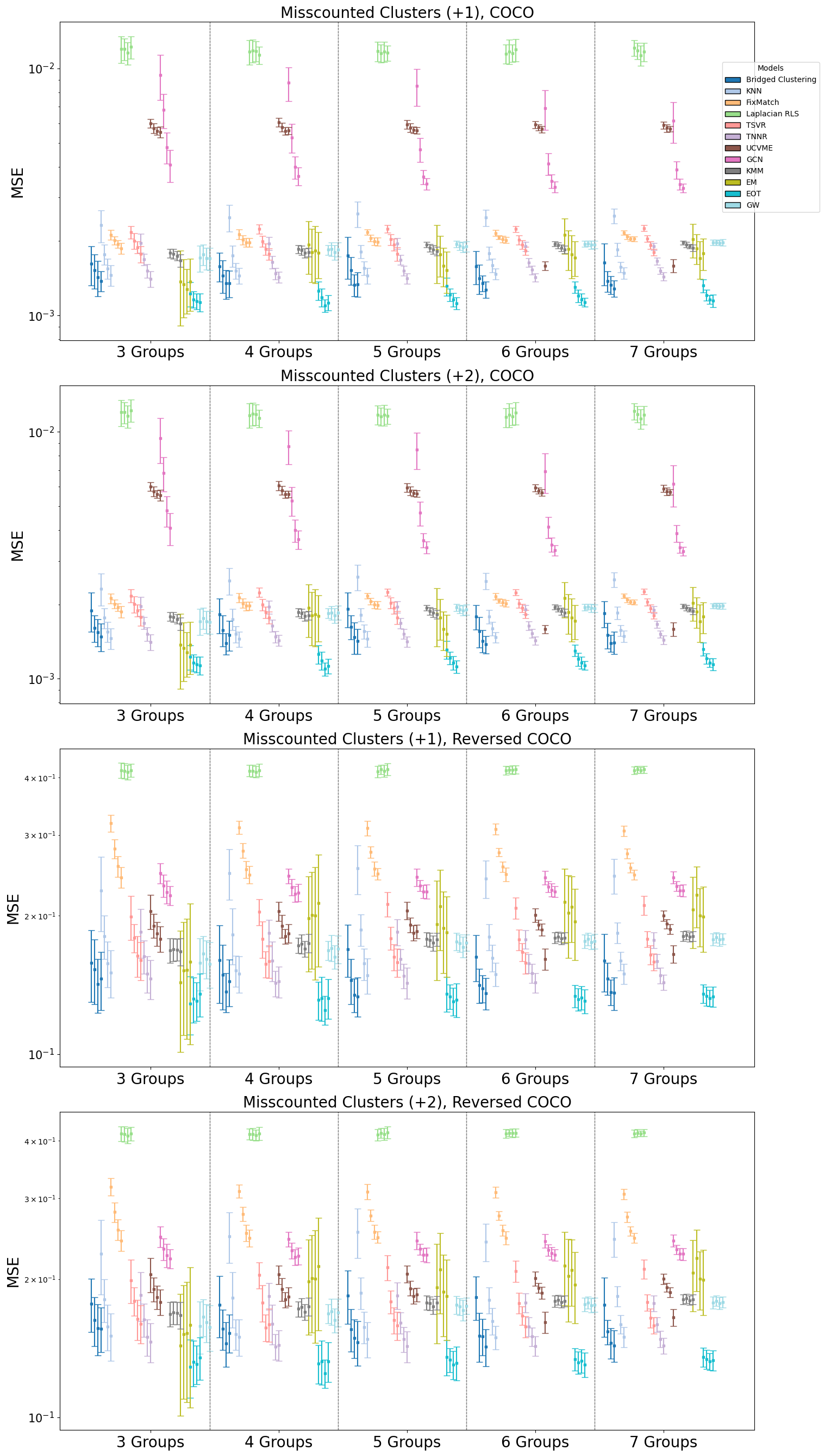}
    \caption{Running main experiment on Misscounted Cluster numbers for COCO and Reversed COCO (Shown inductive results, but transductive outcomes are similar. Miscounting as shown here refers to overcounting the number of clusters, and we note that undercounting experiments also show similar results.) }
    \label{fig:figo}
\end{figure}

\section{Using Different Encoders and Clustering Methods}

\begin{figure}[H]
    \centering
    \includegraphics[width=0.65\linewidth]{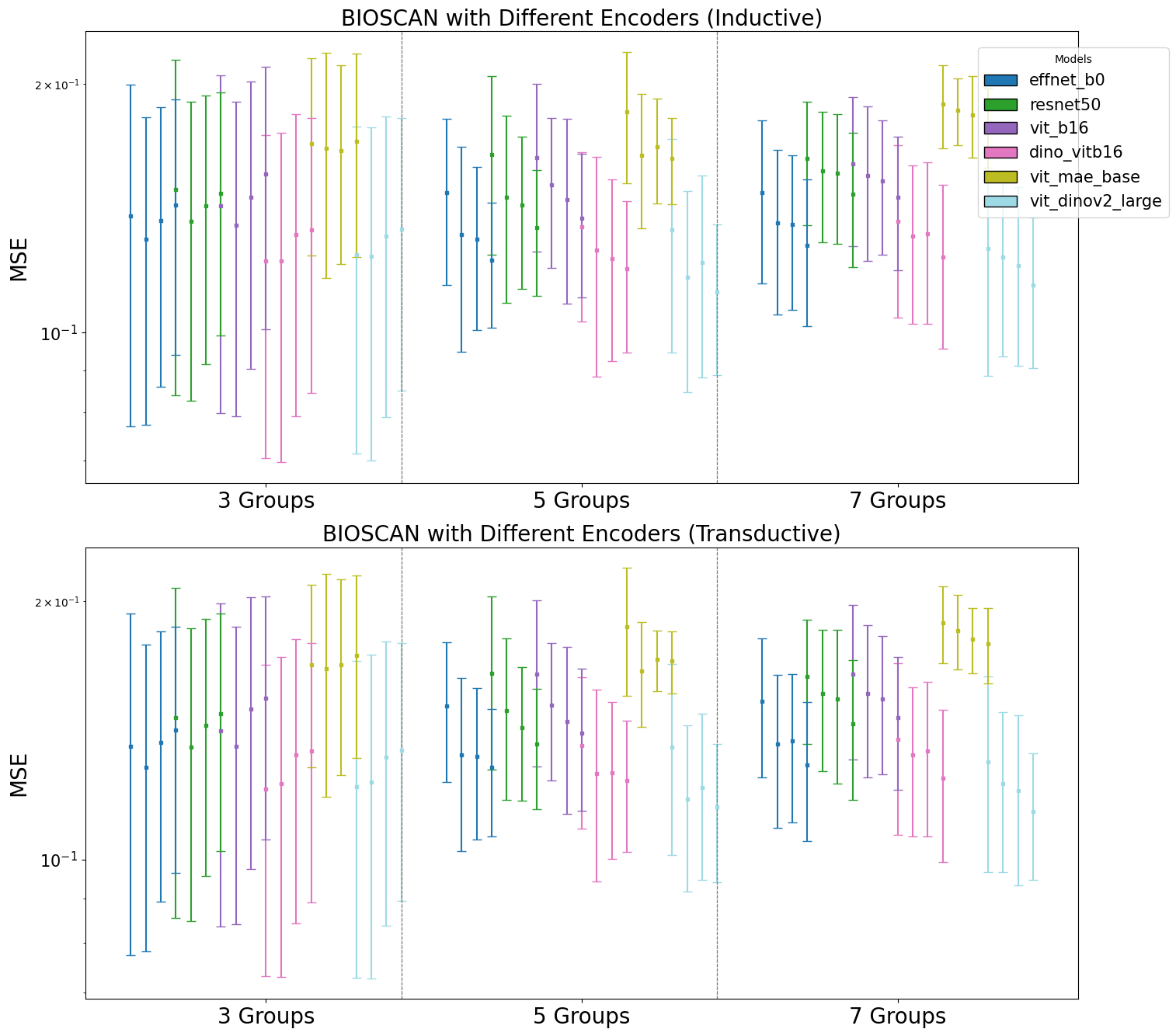}
    \caption{Using Different Image Encoders for BIOSCAN Experiment}
    \label{fig:figp}
\end{figure}

\begin{figure}[H]
    \centering
    \includegraphics[width=0.65\linewidth]{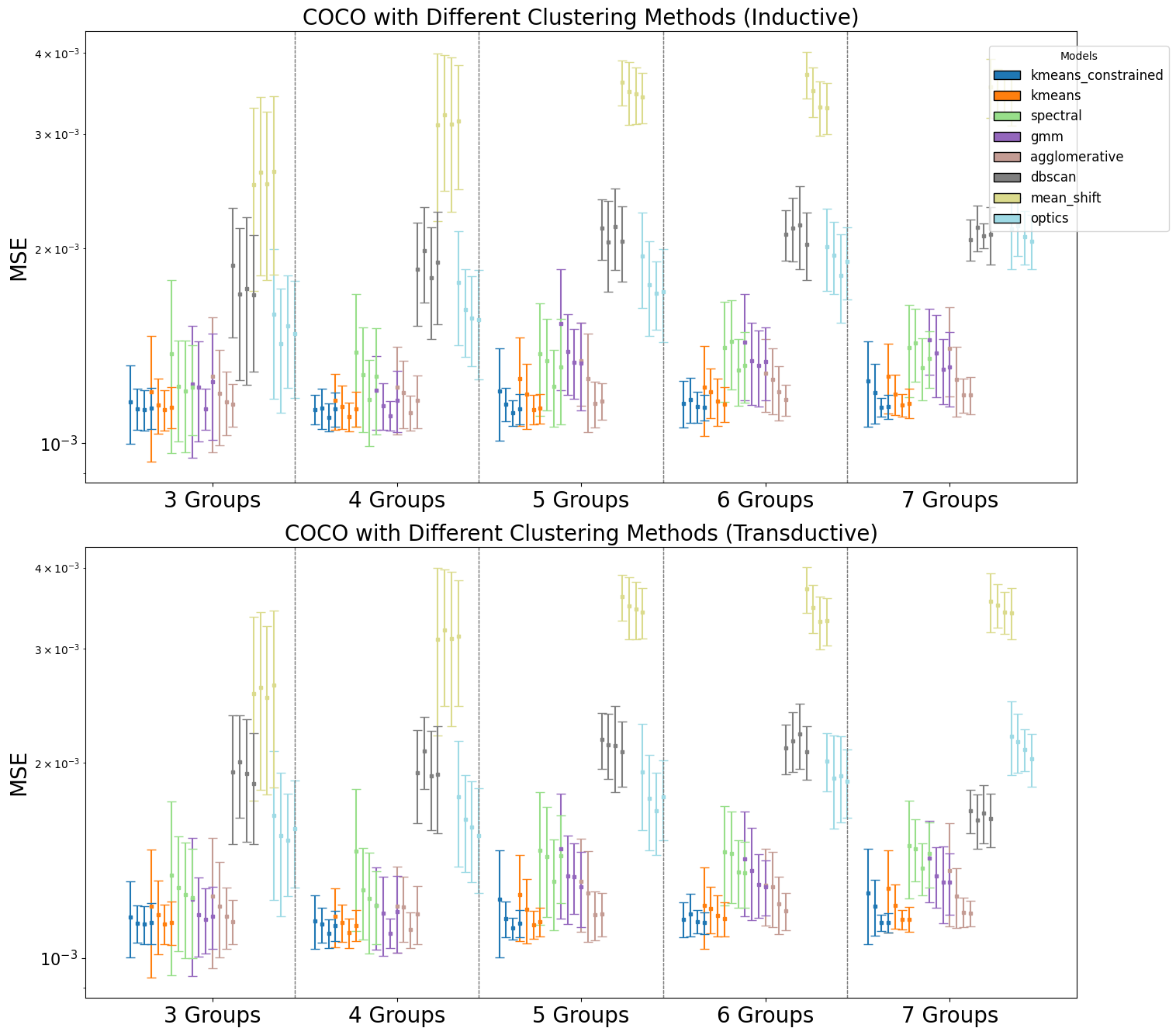}
    \caption{Using Different Clustering Methods for COCO Experiment}
    \label{fig:figq}
\end{figure}

\section{Illustrations for Bridged Clustering}
\label{app:illu}
\begin{figure}[H]
    \centering
    \includegraphics[width=1\linewidth]{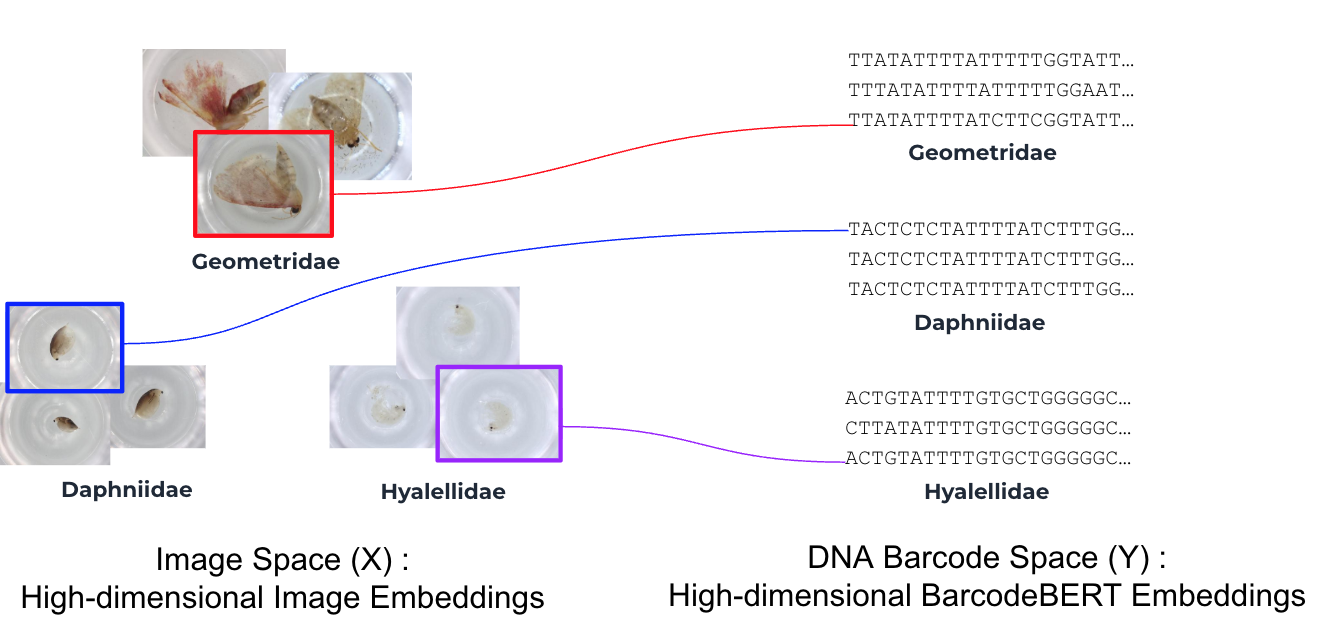}
    \caption{Illustrations for Bridged Clustering in BIOSCAN. In both Image and DNA spaces, datapoints from same species are often clustered together due to embedding proximity. Highlighted datapoints are supervised.}
    \label{fig:figr.png}
\end{figure}

\begin{figure}[H]
        \centering
        \includegraphics[width=1\linewidth]{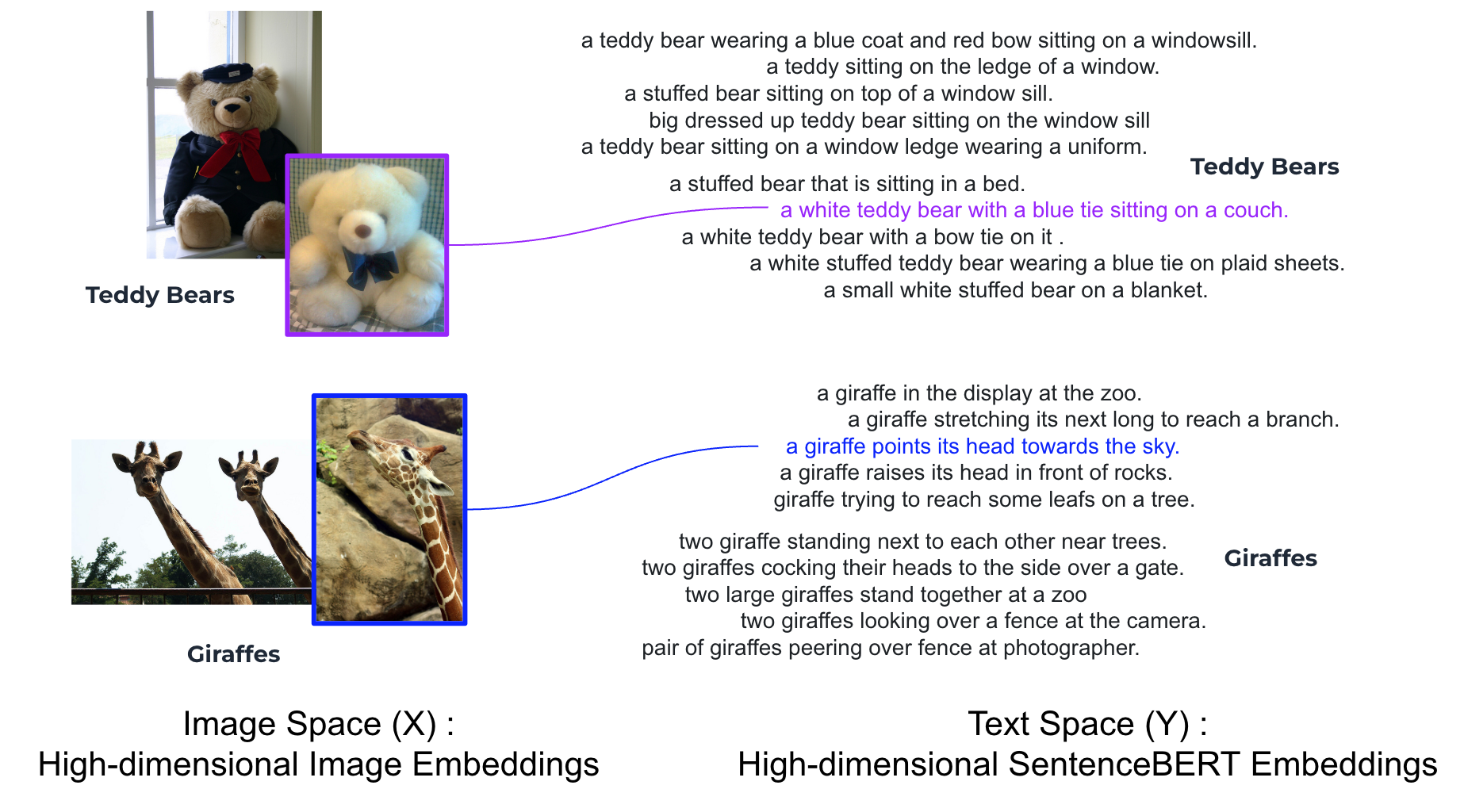}
        \caption{Illustrations for Bridged Clustering in COCO. In both Image and text spaces, datapoints from similar categories are often clustered together due to embedding proximity. Highlighted datapoints are supervised.}
        \label{fig:figA}
\end{figure}

\section{BLEU Score for Image-Captioning Experiments}

\subsection{BLEU Score for Flickr30k Experiments (Higher is better)}
\label{app:bleu}

\begin{figure}[!htbp]
    \centering
    \includegraphics[width=0.65\linewidth]{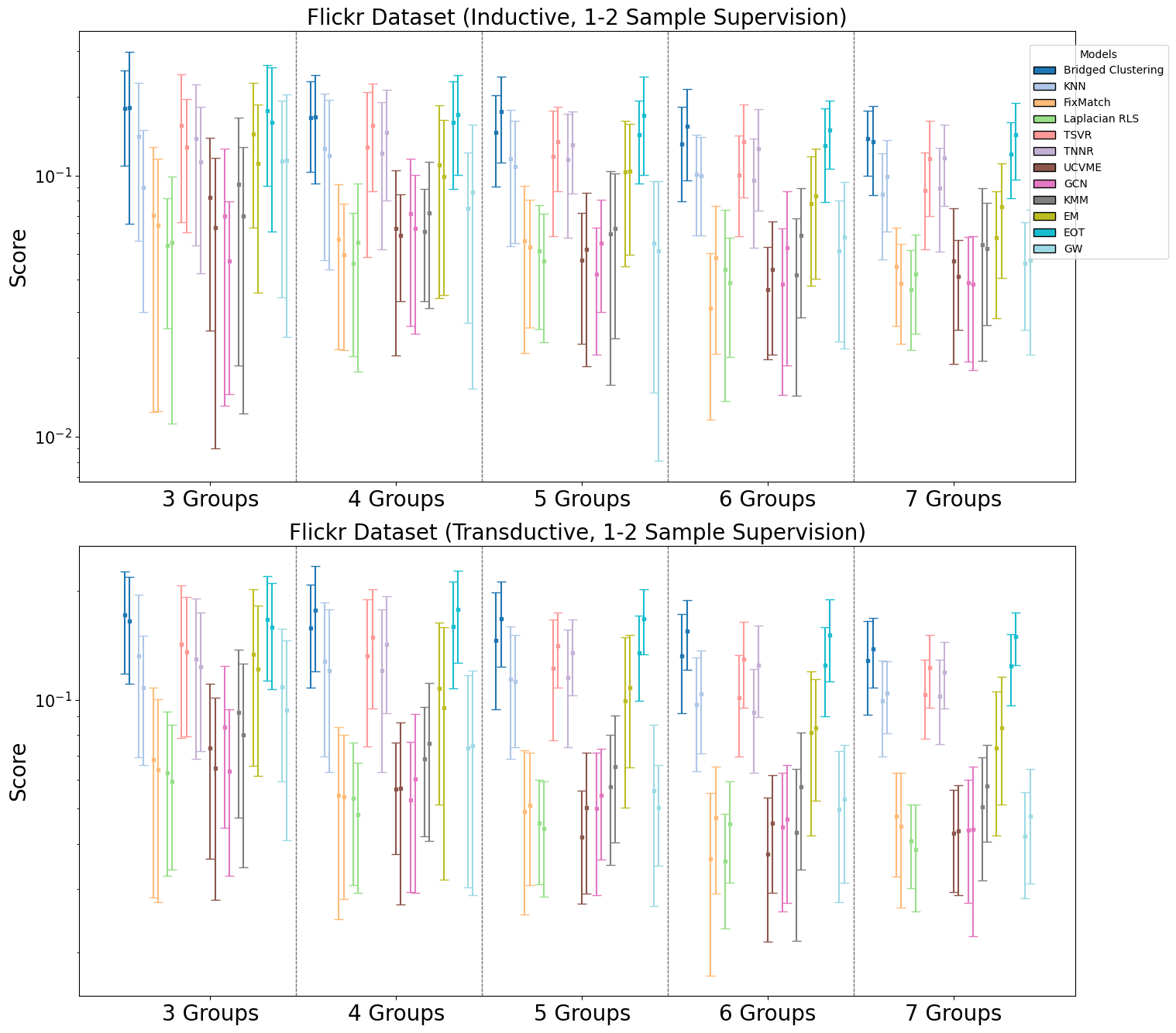}
    \caption{BLEU Score in lower supervision Flickr30k Image-Caption Prediction.}
    \label{fig:figv}
\end{figure}

\begin{figure}[H]
    \centering
    \includegraphics[width=0.65\linewidth]{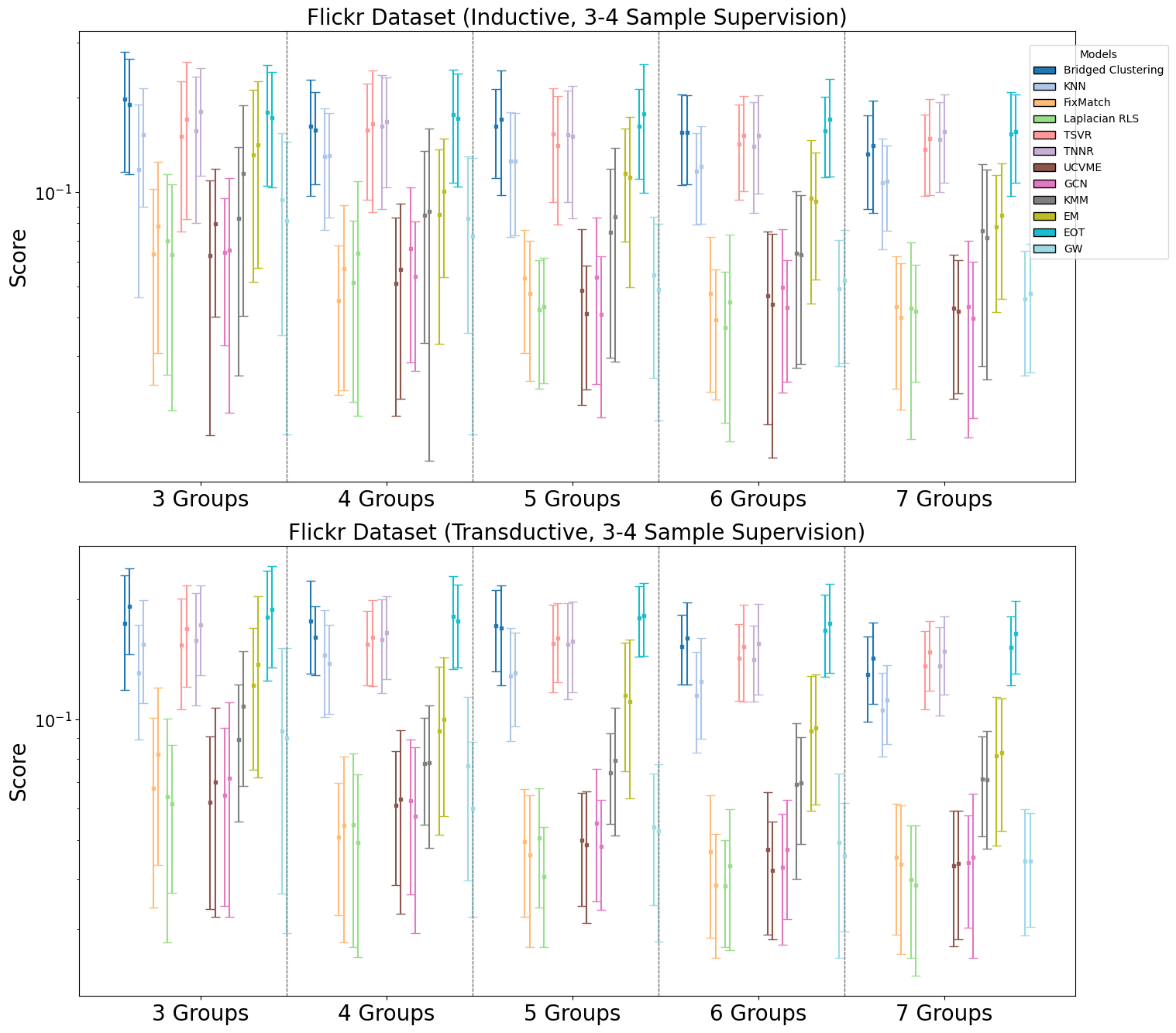}
    \caption{BLEU Score in higher supervision Flickr30k Image-Caption Prediction.}
    \label{fig:figu}
\end{figure}

\subsection{BLEU Score for COCO Experiments (Higher is better)}
\begin{figure}[H]
    \centering
    \includegraphics[width=0.65\linewidth]{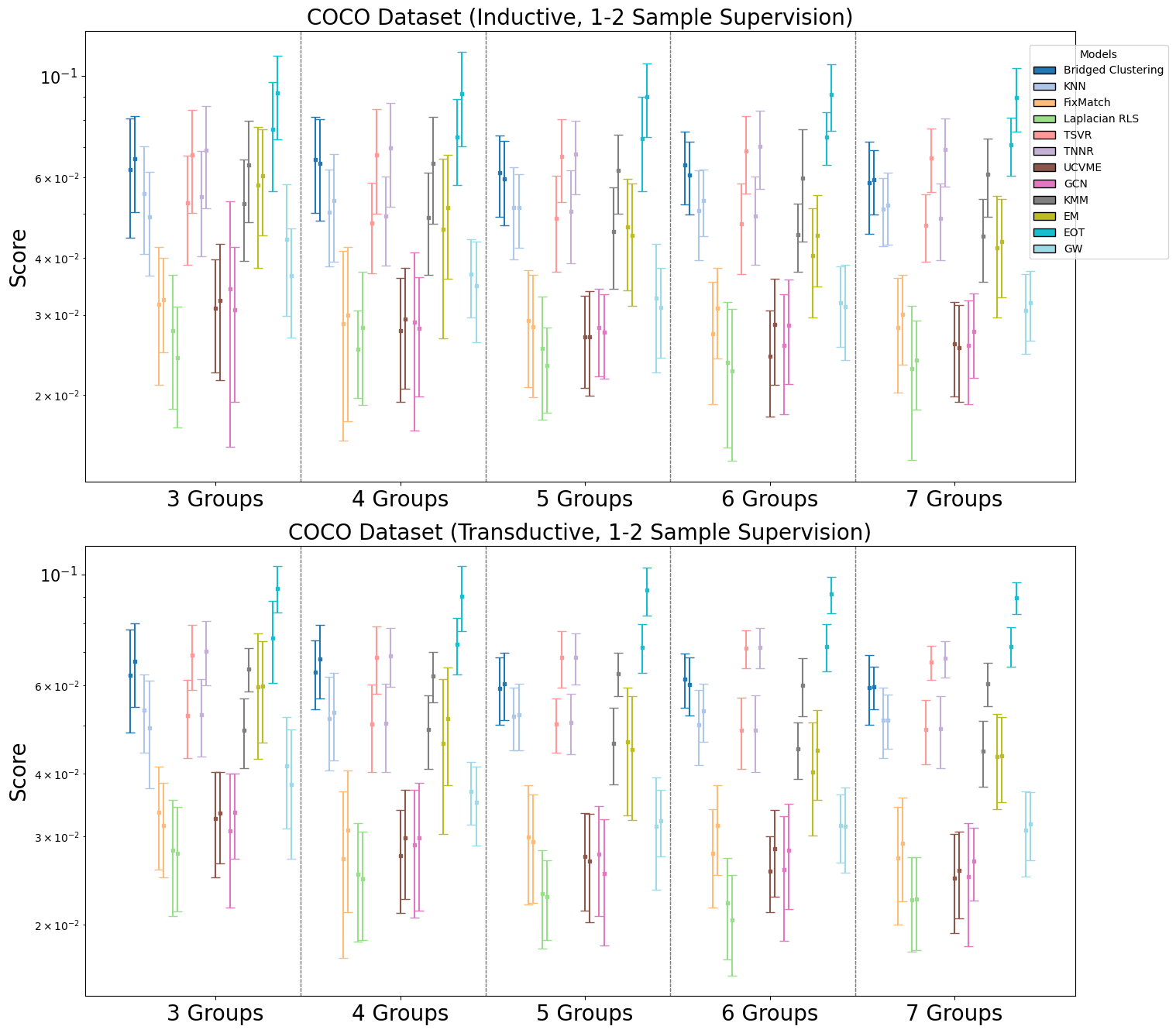}
    \caption{BLEU Score in lower supervision COCO Image-Caption Prediction.}
    \label{fig:figs}
\end{figure}

\begin{figure}[H]
    \centering
    \includegraphics[width=0.65\linewidth]{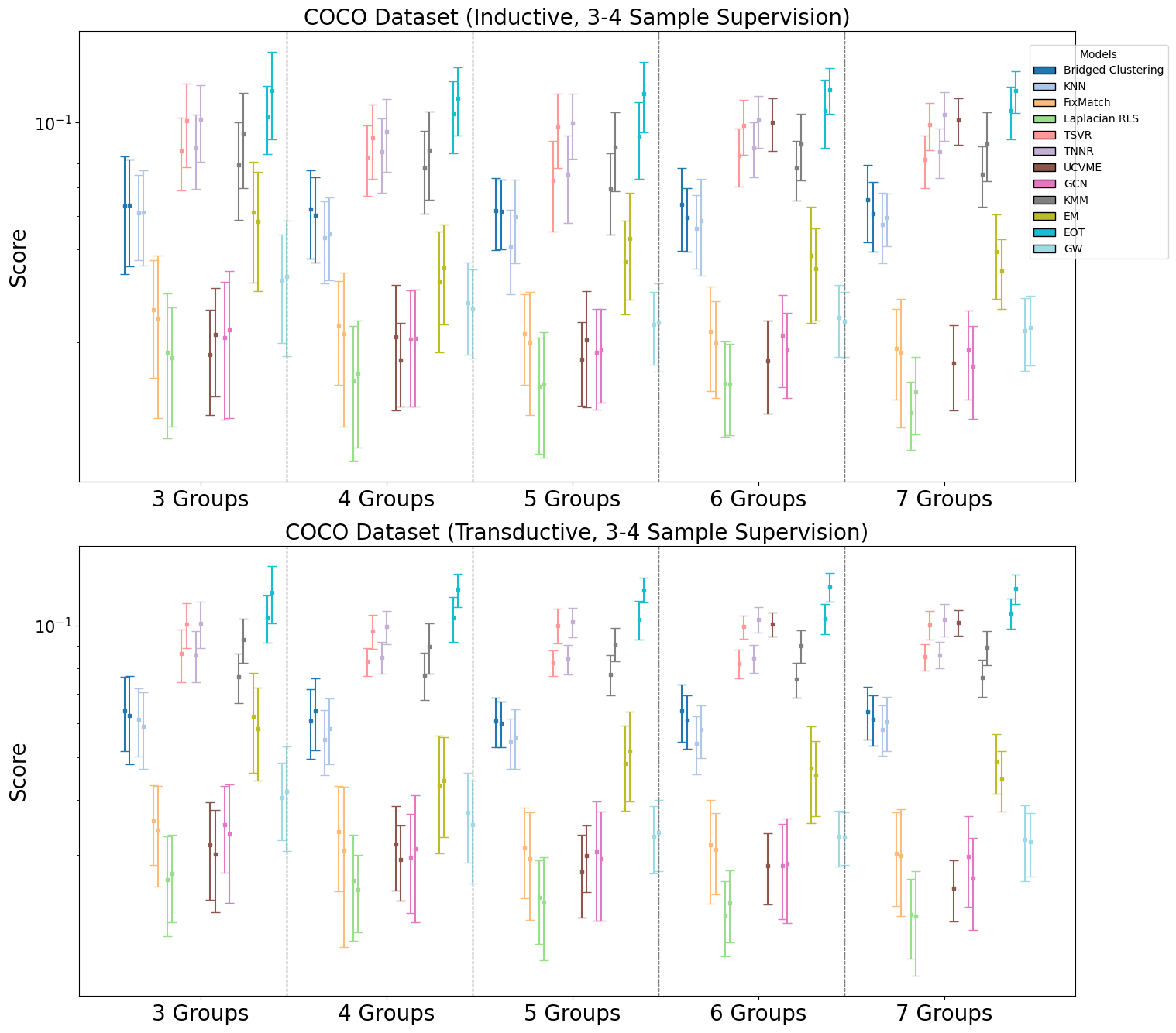}
    \caption{BLEU Score in higher supervision COCO Image-Caption Prediction.}
    \label{fig:figt}
\end{figure}

\subsection{BLEU Score for Wiki Experiments (Higher is better)}

\begin{figure}[H]
    \centering
    \includegraphics[width=0.65\linewidth]{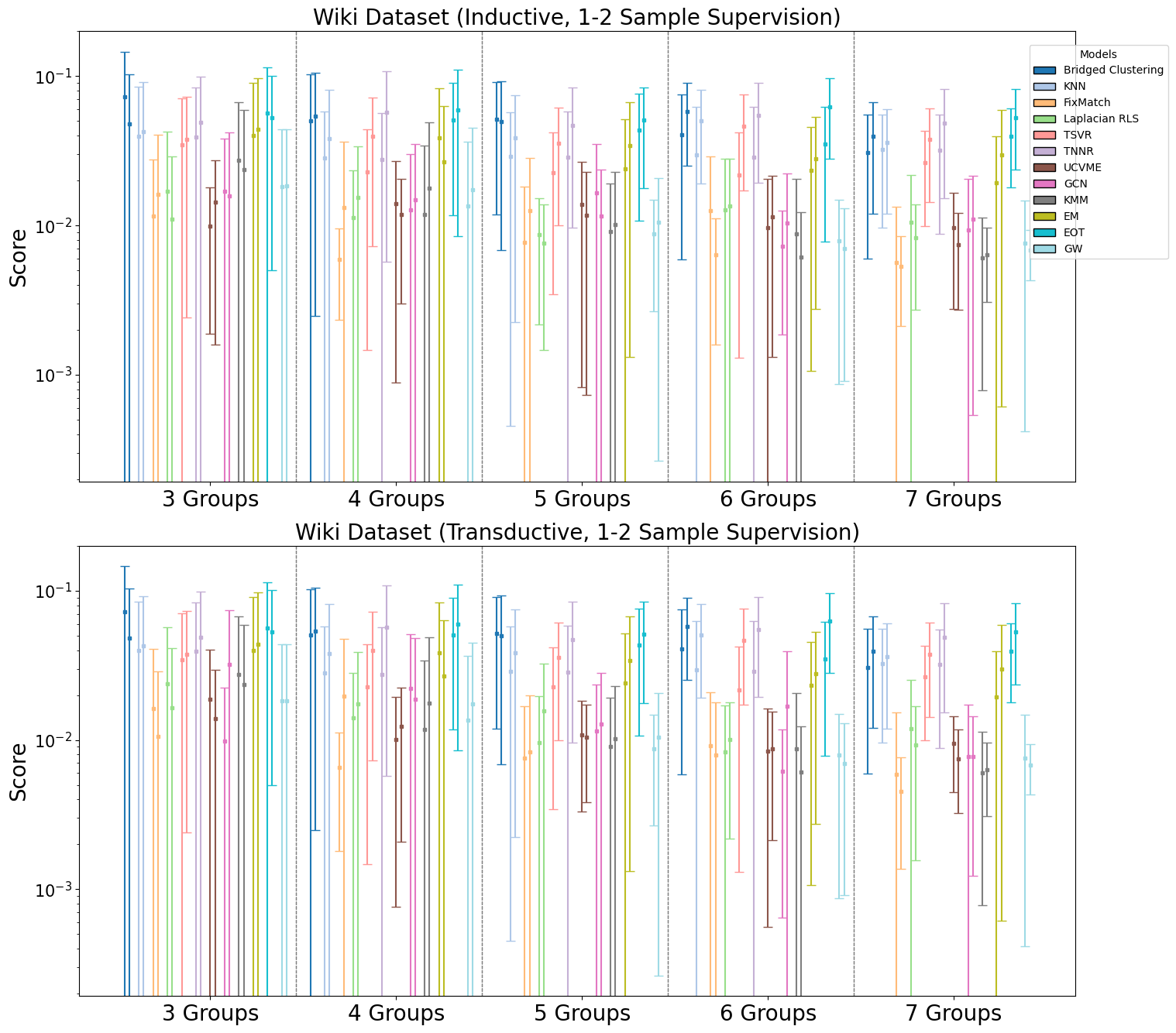}
    \caption{BLEU Score in lower supervision Wiki Image-Caption Prediction.}
    \label{fig:figw}
\end{figure}

\begin{figure}[H]
    \centering
    \includegraphics[width=0.65\linewidth]{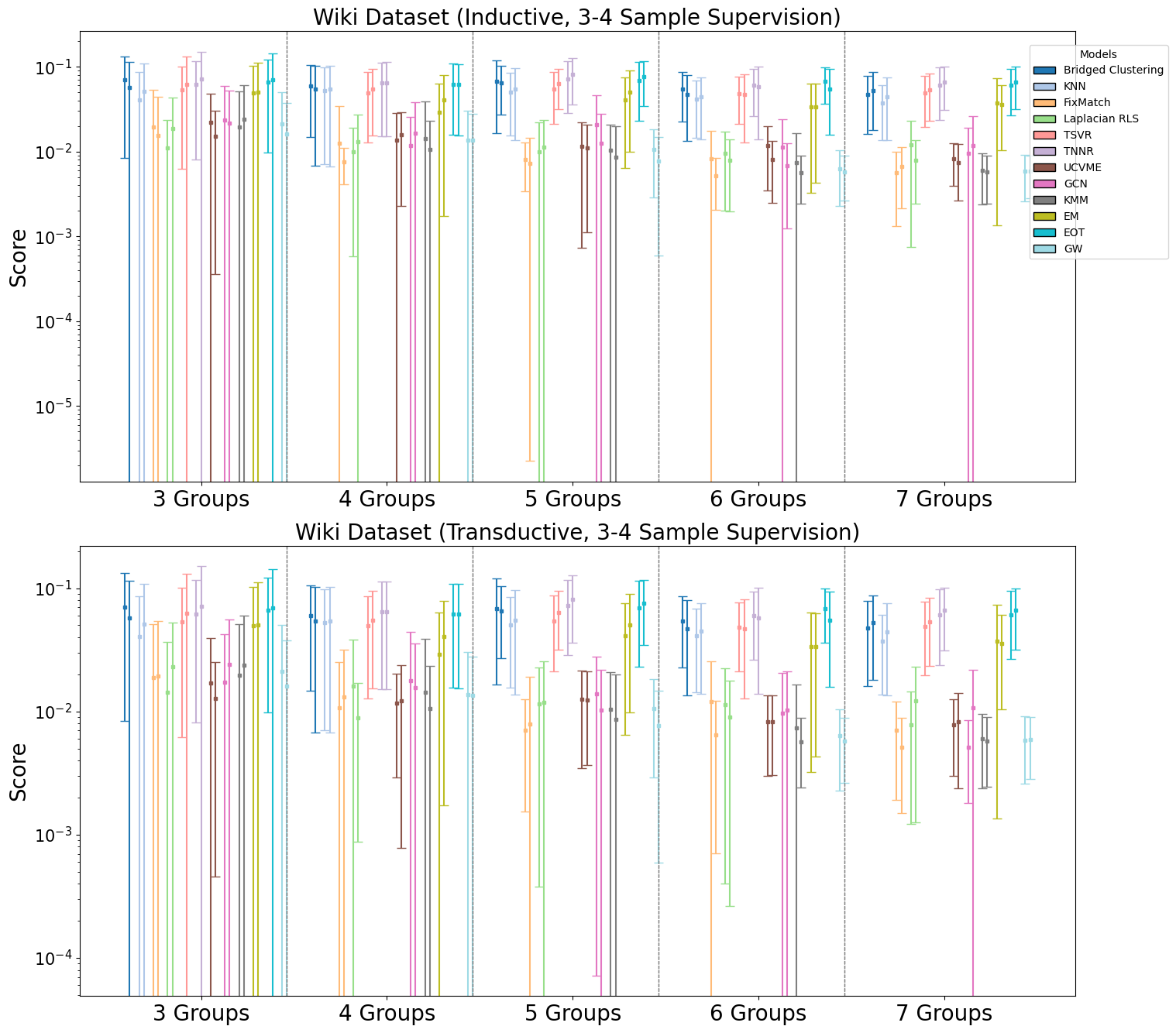}
    \caption{BLEU Score in higher supervision Wiki Image-Caption Prediction.}
    \label{fig:figx}
\end{figure}

\section{LLM Usage}
We used ChatGPT to aid and polish this paper, including grammatical and writing quality checks.

\section{More Information}
For the experiments, we use a machine with 48 cores, 516GB RAM, NVIDIA RTX 6000 Ada Generation 49.1GB; no special hardware is required. 

\end{document}